\documentclass[journal]{IEEEtran}
\usepackage{graphicx}
\usepackage{multirow}
\usepackage{epstopdf}
\usepackage{times}
\usepackage{amsmath}
\usepackage{amssymb}
\usepackage{amsfonts}
\usepackage{color}
\usepackage[colorlinks]{hyperref}
 \usepackage{colortbl}
\usepackage{makecell}
\usepackage{array}
\usepackage{scalerel}
\usepackage{subcaption}
\usepackage{arydshln}
\usepackage[linesnumbered,ruled,vlined,algo2e]{algorithm2e}
\usepackage{algorithmic}
\usepackage{rotating}
\usepackage{adjustbox,lipsum}
\usepackage{dsfont}
\usepackage{booktabs}

\usepackage[table]{xcolor}
\usepackage{pifont}

\newcommand{\ie}{{\em i.e.}}           

\begin{document}

\title{CILP-FGDI: Exploiting Vision-Language Model for Generalizable Person Re-Identification}

\author{Huazhong Zhao,
        Lei Qi,
        Xin Geng

\thanks{The work is supported by NSFC Program (Grants No. 62206052, 62125602, 62076063), China Postdoctoral Science Foundation (Grants No. 2024M750424), Supported by the Postdoctoral Fellowship Program of CPSF (Grant No. GZC20240252), and Jiangsu Funding Program for Excellent Postdoctoral Talent (Grant No. 2024ZB242).}
\thanks{Huazhong Zhao, Lei Qi and Xin Geng are with the School of Computer Science and Engineering, Southeast University, and Key Laboratory of New Generation Artificial Intelligence Technology and Its Interdisciplinary Applications (Southeast University), Ministry of Education, China, 211189 (e-mail: zhaohuazhong@seu.edu.cn; qilei@seu.edu.cn; xgeng@seu.edu.cn).}
\thanks{Corresponding author: Lei Qi.}

}

%
%

\markboth{ }%
{Shell \MakeLowercase{\textit{et al.}}: Bare Demo of IEEEtran.cls for IEEE Journals}

\maketitle

\begin{abstract}
The Visual Language Model, known for its robust cross-modal capabilities, has been extensively applied in various computer vision tasks.
In this paper, we explore the use of CLIP (Contrastive Language-Image Pretraining), a vision-language model pretrained on large-scale image-text pairs to align visual and textual features, for acquiring fine-grained and domain-invariant representations in generalizable person re-identification.
The adaptation of CLIP to the task presents two primary challenges: learning more fine-grained features to enhance discriminative ability, and learning more domain-invariant features to improve the model's generalization capabilities.
To mitigate the first challenge thereby enhance the ability to learn fine-grained features, a three-stage strategy is proposed to boost the accuracy of text descriptions.
Initially, the image encoder is trained to effectively adapt to person re-identification tasks.
In the second stage, the features extracted by the image encoder are used to generate textual descriptions (i.e., prompts) for each image.
Finally, the text encoder with the learned prompts is employed to guide the training of the final image encoder.
To enhance the model's generalization capabilities to unseen domains, a bidirectional guiding method is introduced to learn domain-invariant image features.
Specifically, domain-invariant and domain-relevant prompts are generated, and both positive (i.e., pulling together image features and domain-invariant prompts) and negative (i.e., pushing apart image features and domain-relevant prompts) views are used to train the image encoder.
Collectively, these strategies contribute to the development of an innovative CLIP-based framework for learning fine-grained generalized features in person re-identification.
The effectiveness of the proposed method is validated through a comprehensive series of experiments conducted on multiple benchmarks.
Our code is available at \href{https://github.com/Qi5Lei/CLIP-FGDI}{https://github.com/Qi5Lei/CLIP-FGDI}.

\end{abstract}


\IEEEpeerreviewmaketitle

\section{Introduction}

\IEEEPARstart{G}{eneralizable} person re-identification (DG-ReID), is a vital and challenging facet of computer vision~\cite{Zheng_Yang_Hauptmann_2016, 2020PurifyNet, Ye_Shen_Lin_Xiang_Shao_Hoi_2022, Leng_Ye_Tian_2020,2020Attribute,qi2022label,xie2024prcl,gu2023color}.
This task has found application in various fields such as video surveillance~\cite{wang2021robust}, social media analysis, intelligent transportation systems and so on.
In the context of generalizable person re-identification, this specialized field revolves around the challenging task of recognizing individuals across different camera views or domains \ie, variations in conditions such as lighting, background, and image resolutions), where conditions can vary significantly~\cite{Liao_Shao_2020, Liao_Shao, He_Liu_Liang_Zheng_Liao_Cheng_Mei, Zhang_Dou_Yunlong_Li, xu2022mimic, Liao_Shao_2021,2019Person,2023Dual}.
Thus, the acquisition of models' capability to learn more \emph{fine-grained} and \emph{domain-invariant} feature representations in generalizable person re-identification tasks is of paramount importance.

\begin{figure}[t]
    \centering
    \includegraphics[width=0.95\linewidth]{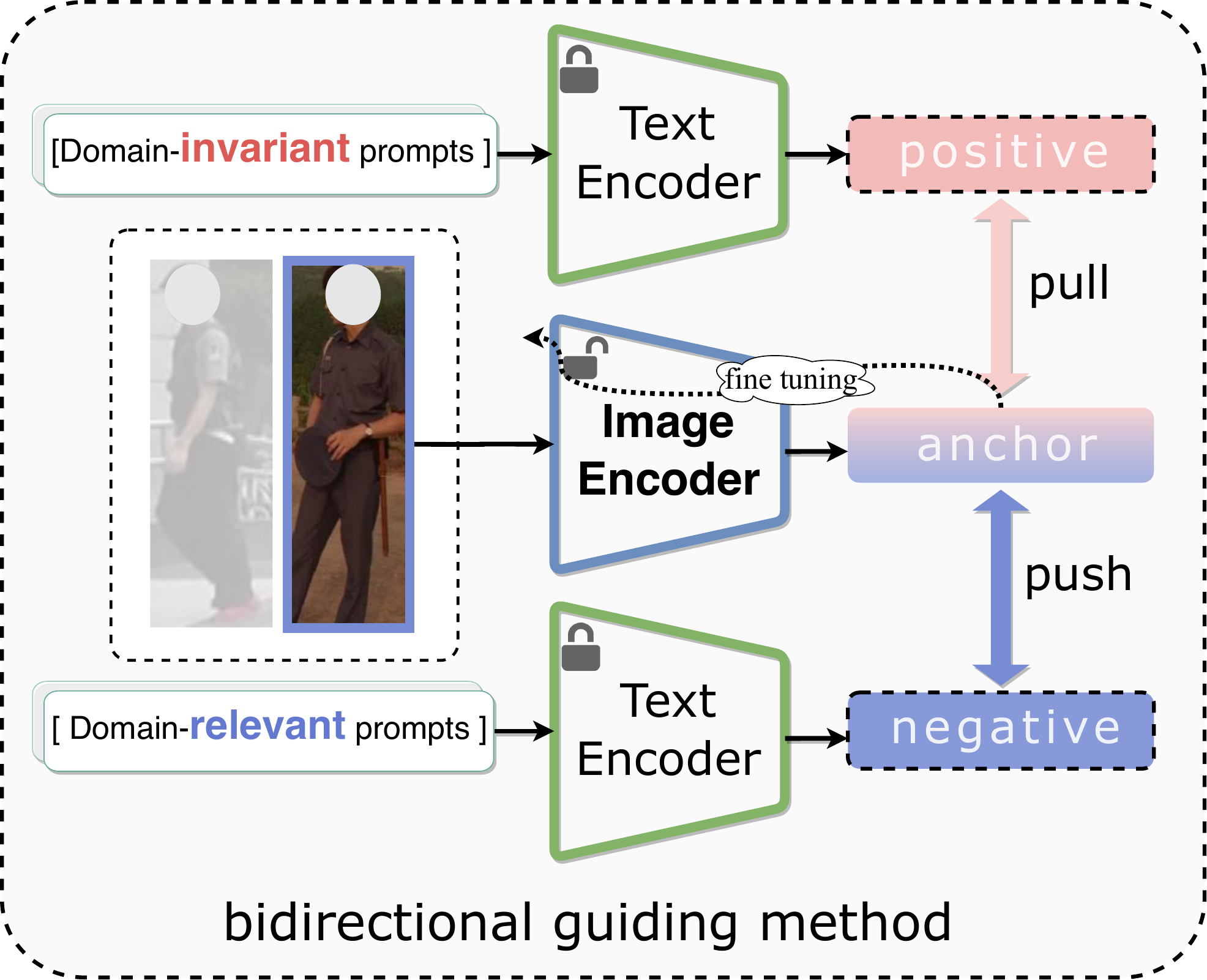}
    \caption{Using the powerful cross-modal capabilities of the Vison-Language model, we employ a bidirectional guiding method, which involves fine-tuning the image encoder using features derived from prompts that are both domain-invariant and domain-relevant. This bidirectional guidance allows the model to learn domain-invariant features effectively.}
  \label{fig:1}

\end{figure}

Pre-trained vision-language models, like CLIP~\cite{radford2021learning}, have recently demonstrated superior performance on various downstream tasks, especially image classification~\cite{Ma_2023_CVPR} and segmentation~\cite{Zhou_2023_CVPR}.
However, there is a lack of work on the application of CLIP to the field of generalizable person re-identification.
To the best of our knowledge, CLIP-ReID~\cite{li2023clip} is a relatively representative work based on CLIP for traditional person re-identification.
CLIP-ReID tackles the challenge of image re-identification (ReID) where labels lack specific text descriptions.
The method introduces a two-stage strategy to leverage CLIP's cross-modal capabilities.
Learnable text tokens for each ID are learned in the first stage, while the second stage fine-tunes the image encoder with static ID-specific text tokens.
By fine-tuning CLIP's visual model, competitive ReID performance is achieved.

In attempting to apply CLIP to this task, we encountered two main challenges: how to enhance the model's  discriminative ability and how to improve the model's generalization capabilities.
In the face of the first challenge, we employ a novel three-stage strategy.
Unlike the two-stage training method used in CLIP-ReID~\cite{li2023clip}, we introduce an initial training stage with a few epochs to allow the image encoder to adapt to the fine-grained task of person re-identification.
This phase significantly aids the subsequent stages in learning more accurate text descriptions, thereby further enhancing the ability to distinguish between different pedestrian IDs.
Subsequently, in the second stage, the features extracted by the image encoder are utilized to generate textual descriptions for each image.
When confronted with the second challenge.
To augment the model's generalization capabilities to unseen domains, we introduce a bidirectional guiding method during the second training stage, which ensures the person features obtained by the image encoder are domain-invariant, as shown in Fig.~\ref{fig:1}.
To be specific, domain-invariant and domain-relevant prompts are learned, and these prompts are employed to assist the image encoder in learning generalized features, considering both positive (\ie, pulling together image features and domain-invariant prompt features) and negative (\ie, pushing apart image features and domain-relevant prompts) perspectives to train the final image encoder.

To provide a thorough understanding of our proposed methodology and to substantiate its efficacy, we delve into detailed explanations of the methods employed.
Furthermore, we present the outcomes of extensive experiments conducted to validate the effectiveness of our method in the subsequent sections.
In summary, our primary contributions can be summarized as follows:
\begin{itemize}
\item To the best of our knowledge, this work is the first to apply CLIP in proposing the CLIP-FGDI framework. CLIP-FGDI, which stands for \emph{CLIP for Fine-Grained and Domain-Invariant feature learning}, effectively harnesses CLIP’s capabilities to extract fine-grained and domain-invariant features, enabling robust performance in generalizable person re-identification.
\item We propose a three-stage learning strategy that fully leverages the cross-modal capabilities of CLIP to accurately describe person features, thereby enhancing the model's discriminative ability.
\item We employ a bidirectional guiding method to constrain image features, ensuring that the features obtained by the image encoder are domain-invariant.
\item Extensive experiments conducted on various standard benchmark datasets demonstrate the significant improvements achieved by our method in the context of generalizable person re-identification tasks.
\end{itemize}

The structure of this paper is outlined as follows: Section~\ref{sec:related_work} provides a literature review on relevant research.
In Section~\ref{sec:method}, we introduce our method.
Section~\ref{sec:experiments} presents the experimental results.
Lastly, we summarize this paper in Section~\ref{sec:conclusion}.

\section{Related Work}
\label{sec:related_work}
In this section, we conduct a thorough investigation of some of the most relevant works, with the aim of delivering a comprehensive overview.

\subsection{Vision-Language Model}
Vision-Language Model often referred to as VLM, is an interdisciplinary field that lies at the intersection of computer vision (CV)~\cite{zhang2023vision} and natural language processing (NLP).
It has obtained significant attention in recent years due to its potential applications in various domains, including image captioning~\cite{hossain2019comprehensive}, visual question answering~\cite{wu2017visual}, cross-modal retrieval~\cite{zhen2019deep} and of course, person re-identification~\cite{li2023clip}.
Early methods predominantly focused on designing feature extraction methods and separate pipelines for visual and textual data.
However, the advent of deep learning has revolutionized the field.
In particular, the introduction of Vision Transformers (ViTs)~\cite{ViT} marked a significant advancement.
Models like CLIP (Contrastive Language-Image Pretraining)~\cite{radford2021learning}, BLIP (Bootstrapping Language-image Pre-training)~\cite{li2022blip,li2023blip}, MiniGPT~\cite{zhu2023minigpt,chen2023minigpt} and so on showcase remarkable capabilities in relating images and text.

Moreover, the field has witnessed the emergence of methods that tackle fine-grained vision-language tasks~\cite{varma2023villa,he2017fine}, which require the model to grasp subtle details and nuanced relationships between images and language.
As vision-language model continues to evolve, it has grown into a dynamic research area, offering a wide range of possibilities and applications.
The development of models that can seamlessly bridge the gap between visual and textual data holds immense promise for the future development of artificial intelligence.

Building upon this, CLIP-ReID~\cite{li2023clip} is a representative work in Vision-Language Models applied to person re-identification.
By leveraging the CLIP model’s contrastive learning framework, CLIP-ReID jointly embeds images and textual descriptions into a shared feature space.
This enables the model to use both visual and textual features to improve person matching across different camera views and environments.

\subsection{Text-to-image Person Re-Identification}
Text-to-image Person Re-Identification focuses on the task of associating textual descriptions with corresponding images to recognize individuals.
This fusion of text and image information holds immense promise for applications in surveillance, security, and visual content retrieval~\cite{jiang2023cross}.

The inception of Text-to-Image Person Re-Identification can be traced back to the broader domain of person re-identification~\cite{zheng2015scalable}, where the objective is to match the same person across different camera views, often under varying conditions.
The integration of text information into this framework signifies a significant shift.
Recent methods in this field aim to bridge the gap between images and textual descriptions.
These methods utilize datasets containing both images and text descriptions of individuals, which serve as valuable resources for training deep neural networks~\cite{li2014deepreid}.
The idea is to map images and text into a shared embedding space, where semantic correspondences can be effectively captured~\cite{li2023clip}.
Text-to-Image Person Re-ID is facilitated by advanced pre-training techniques, inspired by the success of vision-language models.
These models, trained on massive cross-modal datasets, demonstrate superior capabilities in learning fine-grained associations between images and text.
Researchers are increasingly exploring the adaptation of such models to the specific task of person re-identification with textual descriptions~\cite{li2023clip,jiang2023cross,Li_Xiao_Li_Zhou_Yue_Wang_2017,Zhu_Wang_Li_Wan_Jin_Wang_Hu_Hua_2021,ding2021semantically,Wang_Zhu_Xue_Wan_Liu_Wang_Li_2022}.

\subsection{Generalizable Person Re-Identification}
Generalizable Person Re-Identification is a critical research area in computer vision that focuses on the ability of person recognition models to perform well across unseen domains.
In Generalizable Person Re-Identification, one of the fundamental challenges is domain shift, where the source and target domains have significant differences.
So the aim is to reduce the difference in data distributions between the source domains and target domains.

Recent years have seen the development of various methods aimed at addressing this challenge.
For example, the META framework~\cite{xu2022mimic} incorporates an aggregation module to dynamically combine multiple experts through normalization statistics, enabling the model to adapt to the characteristics of the target domain.
Similarly, the ACL framework~\cite{Zhang_Dou_Yunlong_Li} introduces a Cross-Domain Embedding Block (CODE-Block) that explores relationships across diverse domains, ensuring the shared feature space captures both domain-invariant and domain-specific features.
The DFF~\cite{lin2023deep} uses a gradient reversal layer~\cite{ganin2016domain} to enable learning of domain-invariant features, allowing the model to perform well in new situations without requiring additional labeled data.
Moreover, several recent methods utilize Vision Transformers (ViT) for person re-identification.
For instance, TransReID~\cite{He_Luo_Wang_Wang_Li_Jiang_2021}, PAT~\cite{ni2023part}, and DSM+SHS~\cite{Li2023StyleControllableGP}.
In traditional domain generalization tasks using only visual features, learning ``domain-invariant'' features is challenging due to ambiguities caused by factors like background, lighting, or image resolution.

Our proposed method, by incorporating vision-language models, can use text-based prompts corresponding to visual features to guide the learning of ``domain-invariant'' features.
Unlike visual features, textual prompts focus solely on the person, avoiding irrelevant details.
This integration of visual and textual features improves robustness to domain shifts, enhancing generalization across varying environments and effectively addressing challenges like lighting variations, background differences, and different image resolutions.

\section{Method}
\label{sec:method}
\begin{figure*}[t]
  \centering
   \includegraphics[width=0.99\linewidth]{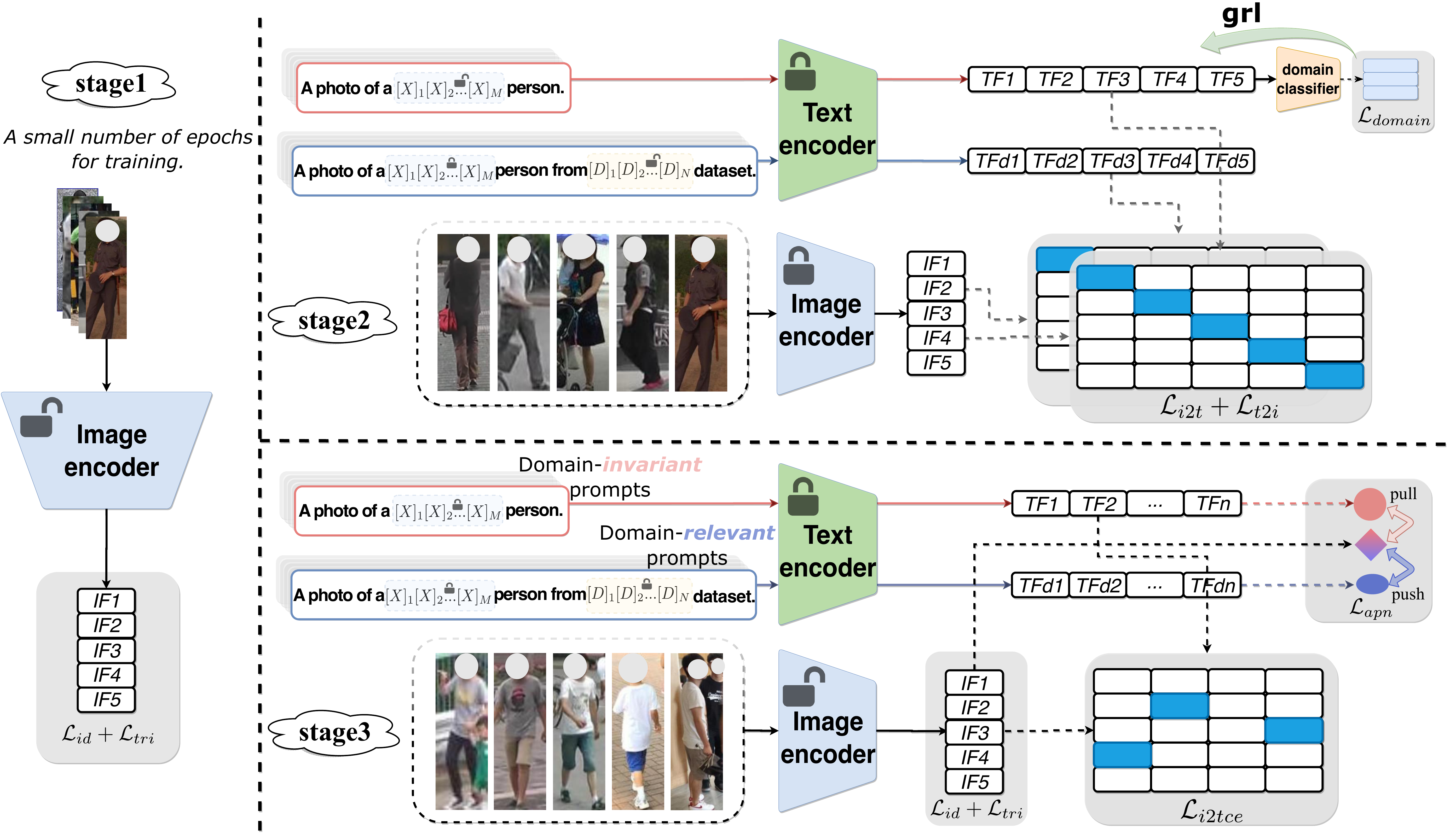}

   \caption{Overview of our method. In the first stage, a small number of epochs are employed to train the image encoder. In the second stage, the Gradient Reversal Layer (GRL) is utilized to learn domain-invariant ID-specific tokens and Domain-specific tokens. In the third stage, fine-tuning of the image encoder is performed using loss designed for downstream tasks. Here, ``IF'', ``TF'' and ``TFd'' represent ``Image Feature'', ``Text Feature'' and ``Text Feature with domain information'', respectively.}
   \label{fig:2}

\end{figure*}

In this section, we present our proposed method base on CLIP for Fine-Grained and Domain-Invariant feature learning (CLIP-FGDI).
The overview of CLIP-FGDI is illustrated in Fig.~\ref{fig:2} and the details are discussed in the following sections.
\subsection{Preliminaries}

\textbf{CLIP} (Contrastive Language–Image Pretraining) is a vision-language pretraining model based on contrastive learning.
It is essential to introduce its core concepts and structure in a format consistent.
Its objective is to embed images and text into a shared embedding space in a manner that brings similar images and text closer to each other within this space.
This approach demonstrate exceptional performance across various computer vision tasks, including image classification, image retrieval, and object detection.
The fundamental idea of CLIP revolves around maximizing the similarity between relevant image-text pairs and minimizing the similarity between irrelevant pairs.
This enables the embeddings for both images and text that exist in a common feature space.

Given a set of image-text pairs \((I, T)\), where \(I\) represents the images and \(T\) represents the corresponding text descriptions, the objective function is designed to maximize the similarity between \(I\) and \(T\) pairs, while minimizing the similarity between negative samples.
This can be formulated as:
\begin{eqnarray}
 \max \sum \log \text{Sim}(I, T) - \sum \log \text{Sim}(I, T') ,
    \label{eq1}
\end{eqnarray}
where \(\text{Sim}(I, T)\) represents a similarity function between images and text descriptions.
The positive pairs \((I, T)\) consist of corresponding image-text pairs, and \(T'\) represents randomly sampled text descriptions that are not associated with the given image \(I\).
CLIP employs large-scale datasets containing images and their textual descriptions to pretrain its model.
During this pretraining phase, the model learns to understand the correspondence between images and text.
Once pretrained, CLIP can be fine-tuned on specific downstream tasks, and its ability to associate images with text descriptions makes it a powerful tool for various applications in computer vision.

\textbf{CLIP-ReID} aims to enhance re-identification (ReID) tasks by pretraining learnable text tokens.
The model undergoes a two-stage training process.

The first stage introduces ID-specific text tokens, $[X]_m$, and optimizes them. This phase focuses on ambiguous text description learning while keeping image and text encoders fixed.
It uses $L_{i2t}$ and $L_{t2i}$ loss functions similar to CLIP but modifies $L_{t2i}$ to account for multiple positive samples:
\begin{eqnarray}
\mathcal{L}_{t2i}(y_i) = -\frac{1}{|P(y_i)|}\sum_{p\in P(y_i)}\log\frac{e^{s(V_p, T_{y_i})}}{\sum_{a=1}^B e^{s(V_a, T_{y_i})}} ,
        \label{eq2}
\end{eqnarray}
and the loss function $L_{\text{stage1}}$ is defined as:
\begin{eqnarray}
\mathcal{L}_{\text{stage1}} = \mathcal{L}_{i2t} + \mathcal{L}_{t2i} .
        \label{eq3}
\end{eqnarray}

The second stage only optimize the parameters of the image encoder.
CLIP-ReID employs the most commonly used triplet loss $L_{tri}$ and ID loss $L_{id}$ with label smoothing for optimization.
Moreover, the image-to-text cross-entropy loss $L_{i2tce}$ is also used, calculated as:
\begin{eqnarray}
\mathcal{L}_{i2tce}(i) = -\sum_{k=1}^N q_k\log\frac{e^{s(V_i, T_{y_k})}}{\sum_{a=1}^N e^{s(V_i, T_{y_a})}} ,
        \label{eq4}
\end{eqnarray}
and the loss for the second training stage is defined as:
\begin{eqnarray}
\mathcal{L}_{\text{stage2}} = \mathcal{L}_{id} + \mathcal{L}_{tri} + \mathcal{L}_{i2tce} .
\label{eq5}
\end{eqnarray}

\subsection{CLIP-FGDI}

When CLIP-ReID is applied to Generalizable Person Re-Identification, it encounters two key challenges.
First, directly introducing ID-specific learnable tokens to learn ambiguous text descriptions raises concerns about whether these tokens can accurately describe individuals, especially when the image encoder has not yet encountered the downstream task (ReID).
Second, the common issue of domain shift in cross-domain person re-identification often leads to a performance drop.
To mitigate the first challenge, we adopt a three-stage learning strategy.
In the initial phase, we initialize the learning of the image encoder.
This process aims to ensure the model effectively captures fine-grained features, leading to more precise text descriptions and further enhancing the image encoder's ability to extract more discriminative features.
For the second challenge, a novel loss function is proposed in the second and third stages to achieve bidirectional guidance, where text-based features are used to guide the image encoder in learning domain-invariant features.
This approach effectively mitigates domain shift issues, as illustrated in Fig.~\ref{fig:1}.
Next, we will provide a detailed introduction to these three stages, along with the specific learning objectives for each stage.

\textbf{The first stage.}
Unlike CLIP-ReID, our method begins with the first stage focused on fine-tuning the image encoder. This stage aims to endow the image encoder with the capability to extract fine-grained features specific to the task of person re-identification. Enhancing the image encoder in this manner is crucial for the subsequent learning of ID-specific tokens, thereby increasing the model's reliability.
While adding stages may increase the training cost, this cost is minimal as we only require a few additional epochs to achieve substantial improvements in results.
In our experiments, we set the number of epochs for the first stage to 3, incurring only a minor increase in training expenses.

\textbf{The Second Stage. }
As shown in Fig.~\ref{fig:2}, this stage focuses on learning ID-specific tokens. During this stage, both the text encoder and image encoder parameters are fixed. It is important to note that the final text feature passes through a domain classifier, and we use the domain loss, \(\mathcal{L}_{\text{domain}}\), to constrain it. The domain loss is represented as follows:
\begin{eqnarray}
    \mathcal{L}_{\text{domain}} = -\sum_{j=1}^N y_j \log(p_j),
    \label{eq6}
\end{eqnarray}
where \(N\) is the number of domain classes, \(y_j\) is the true label, and \(p_j\) is the predicted probability for the \(j\)-th class. During backpropagation, we apply the Gradient Reversal Layer (GRL) to perform gradient reversal, ensuring that the ID-specific tokens contain as few domain-specific features as possible. The overall loss for this stage is:
\begin{eqnarray}
\mathcal{L}_{\text{stage2}} = \mathcal{L}_{i2t} + \mathcal{L}_{t2i} + \alpha \mathcal{L}_{\text{domain}},
\label{eq7}
\end{eqnarray}
where \(\mathcal{L}_{i2t}\) is the image-to-text loss, \(\mathcal{L}_{t2i}\) is the text-to-image loss, and \(\alpha\) is a hyperparameter set to 0.01 in our experiments.

The operations described above are sufficient to promote the image encoder's ability to learn domain-invariant features for the next stage.
To further enhance the image encoder's cross-domain capabilities, we also train Domain-specific tokens after learning the ID-specific tokens. During the learning of Domain-specific tokens, the parameters of the text encoder, image encoder, and ID-specific tokens remain fixed.

Additionally, the text prompt design is structured as follows: ``A photo of a \([X]_1[X]_2[X]_3\ldots[X]_M\) person from \([D]_1[D]_2[D]_3\ldots[D]_N\) dataset.'', where \(M\) is set to 4~\cite{li2023clip}, based on the research findings of CLIP-ReID. In our experiments, the source domain dataset comprises at most four instances.
We believe a single prompt token is fully capable of learning domain descriptions, so we set \(N\) to 1.

\textbf{The Third Stage.}
In the third stage, we introduce two text prompts: one without Domain-specific tokens and one with Domain-specific tokens.
To further encourage the model to learn domain-invariant features while avoiding domain-relevant features, we introduce a new triplet loss in addition to the previously defined losses:
\begin{eqnarray}
\mathcal{L}_{apn} = \max(0, \text{Sim}(IF_{a}, TF_{p}) - \text{Sim}(IF_{a}, TF_{n}) + m),
\label{eq8}
\end{eqnarray}
where \(IF_{a}\) represents the output feature of the image encoder, \(TF_{p}\) denotes the output feature of the prompt without Domain-specific tokens, \(TF_{n}\) signifies the output feature of the prompt with Domain-specific tokens, and \(m\) is a margin parameter, set to 0.3, a commonly used value in related works~\cite{He_Luo_Wang_Wang_Li_Jiang_2021, li2023clip}.

The inclusion of this new triplet loss, \(\mathcal{L}_{apn}\), aims to enhance the learning of domain-invariant features during this stage of the training process. The design of this loss function is to pull together image features and domain-invariant prompt features while simultaneously pushing apart image features and domain-relevant prompt features.
The overall loss for the third training stage is defined as:
\begin{eqnarray}
\mathcal{L}_{\text{stage3}} = \mathcal{L}_{id} + \mathcal{L}_{tri} + (1-\beta)\mathcal{L}_{i2tce} + \beta \mathcal{L}_{apn},
\label{eq9}
\end{eqnarray}
where \(\mathcal{L}_{id}\) is the identity loss, \(\mathcal{L}_{tri}\) is the triplet loss, \(\mathcal{L}_{i2tce}\) is the image-to-text cross-entropy loss, and \(\beta\) is a hyperparameter set to 0.3.

By combining these losses, the third stage aims to balance the learning of domain-invariant features and the suppression of domain-relevant features, further enhancing the robustness and generalization capability of the model in person re-identification tasks.

\textbf{Remark 1.}
In \underline{Stage 1}, the image encoder is trained using labeled image data from a person re-identification dataset, where the images are preprocessed and fed into the encoder without involving textual data. In \underline{Stage 2}, textual descriptions (prompts) are generated based on the image features extracted in the first stage, with the image encoder's learned features guiding the learning of these prompts. Finally, in \underline{Stage 3}, both image and text features are combined, where the positive and negative prompts generated in Stage 2 guide the image encoder, enhancing its performance. This multi-stage process enables the model to learn both fine-grained and domain-invariant features by leveraging both image and text data.

\textbf{Remark 2.}
The BGM can be understood as an idea to achieve domain generalization during the second and third stages by aligning image features with domain-invariant text features and distancing them from domain-relevant text features.
And the \(\mathcal{L}_{apn}\) loss function implements this idea.
Without \(\mathcal{L}_{apn}\), the model only aligns image features with domain-invariant text features by contrastive loss.

\begin{table*}[t]
\centering
\caption{Comparison with state-of-the-art methods under Protocol-1. \textbf{Bold} indicates the best result and \underline{underline} represents the second best result. The subscripts and superscripts on the numbers represent the endpoints of the 95\% confidence interval and the ``*'' indicates results obtained based on the open-source code.}

\begin{tabular}{c|c|l|c||c c|c c|c c|c c|c c}
    \Xhline{1px}
    \multirow{2}*{~Setting~}& \multirow{2}[0]*{Backbone} & \multicolumn{1}{c|}{\multirow{2}*{Method}} &\multirow{2}*{Reference} & \multicolumn{2}{c|}{PRID} & \multicolumn{2}{c|}{GRID} & \multicolumn{2}{c|}{VIPeR} & \multicolumn{2}{c|}{iLIDs} & \multicolumn{2}{c}{Average} \\
    \cline{5-14}
      &  &  & & mAP & R1 & mAP & R1 & mAP & R1 & mAP & R1 & ~~~mAP~~~ & ~~~R1~~~\\
    \hline
    \hline
    \multirow{12}[0]{*}{~Protocol-1~}  & \multirow{7}[0]{*}{CNN}& QAConv\(_{50}\)& ECCV\(_{2020}\) & 62.2 & 52.3 & 57.4 & 48.6 & 66.3 & 57.0 & 81.9 & 75.0 & 67.0 & 58.2\\

     & & M\(^3\)L & CVPR\(_{2021}\) & 65.3 & 55.0 & 50.5 & 40.0 & 68.2 & 60.8 & 74.3 & 65.0 & 64.6 & 55.2\\

     & & MetaBIN &CVPR\(_{2021}\) & 70.8 & 61.2 & 57.9 & 50.2 & 64.3 & 55.9 & 82.7 & 74.7 & 68.9 & 60.5\\
     & & META &ECCV\(_{2022}\) & 71.7 & 61.9 & 60.1 & 52.4 & 68.4 & 61.5 & 83.5 & 79.2 & 70.9 & 63.8\\
     & & ACL & ECCV\(_{2022}\) &73.4 & 63.0 & \underline{65.7} & 55.2 & 75.1 & 66.4 & \underline{86.5} & \underline{81.8} & \underline{75.2} & \underline{66.6}\\

     &   &   ReFID & TOMM\(_{2024}\) & 71.3 & 63.2 & 59.8  & \underline{56.1} & 68.7 & 60.9 &  84.6 & 81.0 & 71.1 & 65.3 \\
    &   &  GMN & TCSVT\(_{2024}\) & 75.4 &66.0& 64.8 &54.4& \underline{77.7}& \underline{69.0} & - & - & - & -  \\

    \cline{2-14}

    & \multirow{6}[0]{*}{ViT}&  ViT-B\(^*\) & ICLR\(_{2021}\) &63.8	&52.0	&56.0	&44.8	&74.8	&65.8	&76.2	&65.0	&67.7	&56.9\\
        &  &  TransReID\(^*\)  & ICCV\(_{2021}\) &68.1 &	59.0 &	60.8  &	49.6 &	69.5&60.1 &	79.8 & 68.3 & 69.6 & 59.3 \\
    & &  CLIP-ReID\(^*\)  & AAAI\(_{2023}\)  & 68.3 &	57.0 &	58.2 &	48.8	 &69.3 &	60.1 &	83.4	 &75.0 &	69.8	 &60.2\\
& &  PAT\(^*\)  & ICCV\(_{2023}\)  &57.9	&46.0 &	54.5&	45.6&	67.8	&60.1	&78.1	&66.7	&64.6	& 54.6\\
     & &  DSM+SHS  & MM\(_{2023}\)  & \underline{78.1} &  \underline{69.7} & 62.1& 53.4 & 71.2&62.8  & 84.8 & 77.8  & 74.1  &66.0\\
      & &  \textbf{Ours} & This paper & \textbf{80.1} & \textbf{71.0} &\textbf{75.1} & \textbf{66.7} & \textbf{81.9} & \textbf{76.0} & \textbf{91.8} & \textbf{88.3} & \textbf{~~82.2\(_{\scaleto{{80.8}}{3.0pt}} ^{\scaleto{{83.6}}{3.0pt}}\)} & \textbf{~~75.5\(_{\scaleto{{73.8}}{3.0pt}} ^{\scaleto{{74.4}}{3.0pt}}\)}\\

    \Xhline{1px}
\end{tabular}

\label{tab:1}

\end{table*}

\section{Experiments}
\label{sec:experiments}

\subsection{Experimental Settings}
\textbf{Datasets.}
We conduct extensive experiments on serval public ReID datasets, which include Market1501~\cite{Zheng_Shen_Tian_Wang_Wang_Tian_2015}, MSMT17~\cite{Wei_Zhang_Gao_Tian_2018}, CUHK02~\cite{Li_Wang_2013}, CUHK03~\cite{li2014deepreid}, CUHK-SYSU~\cite{Xiao_Li_Wang_Lin_Wang_2016}, PRID~\cite{Hirzer_Beleznai_Roth_Bischof_2011}, GRID~\cite{Loy_Xiang_Gong_2010}, VIPeR~\cite{Gray_Tao_2008}, and iLIDs~\cite{Zheng_Gong_Xiang_2009}.
In the interest of brevity, we employ abbreviations to denote these datasets, with Market1501 represented as M, MSMT17 as MS, CUHK02 as C2, CUHK03 as C3, and CUHK-SYSU as CS.
To assess the effectiveness of our method, we utilize two widely recognized evaluation metrics, namely the Cumulative Matching Characteristics (CMC) and the mean average precision (mAP).
These metrics are extensively used in the evaluation of person re-identification models.

\textbf{Experimental protocols.}
In the field of generalizable person re-identification, researchers frequently employ three distinct experimental protocols.
The first, known as Protocol-1, entails training the model on datasets like Market1501, CUHK02, CUHK03, and CUHK-SYSU, and subsequently evaluating its performance on datasets such as PRID, GRID, VIPeR, and iLIDs.
The second protocol, termed Protocol-2, revolves around a single-domain testing approach.
The model is exclusively tested using one dataset, which could be Market1501, MSMT17, CUHK-SYSU, or CUHK03, and the remaining datasets are utilized for model training.
The third protocol, Protocol-3, closely resembles Protocol-2, differing primarily in whether both training and testing data from the source domains are used to train the model.
These standardized protocols offer a framework for assessing the generalizability of models across a diverse range of domains.

\begin{table*}[ht]
  \centering
  \caption{Comparison with state-of-the-art methods under Protocol-2 and Protocol-3. \textbf{Bold} indicates the best result and \underline{underline} represents the second best result. The subscripts and superscripts on the numbers represent the endpoints of the 95\% confidence interval and the ``*'' indicates results obtained based on the open-source code.}

  \begin{tabular}{c|c|l|c||c c|c c|c c|c c}
        \Xhline{1px}
    \multirow{2}[0]{*}{Setting} &\multirow{2}[0]*{Backbone}  &  \multicolumn{1}{c|}{\multirow{2}*{Method}} &\multirow{2}*{Reference}& \multicolumn{2}{c|}{~~M+MS+CS\(\rightarrow\)C3~~} & \multicolumn{2}{c|}{~~M+CS+C3\(\rightarrow\)MS~~} & \multicolumn{2}{c|}{~~MS+CS+C3\(\rightarrow\)M~~} & \multicolumn{2}{c}{Average} \\
        \cline{5-12}
        &   &    &  &mAP   & R1    &mAP   & R1    &mAP   & R1    &mAP   &R1 \\
    \hline
    \hline

          \multirow{12}[0]{*}{Protocol-2}  &\multirow{7}[0]{*}{CNN}& QAConv\(_{50}\) &~~ECCV\(_{2020}\)~~ & 25.4  & 24.8  & 16.4  & 45.3  & 63.1  & 83.7  & 35.0    & 51.3 \\
          &  &M3L  &CVPR\(_{2021}\)& 34.2  & 34.4  & 16.7  & 37.5  & 61.5  & 82.3  & 37.5  & 51.4 \\
          & & MetaBIN &CVPR\(_{2021}\) & 28.8  & 28.1  & 17.8  & 40.2  & 57.9  & 80.1  & 34.8  & 49.5 \\
          & & META & ECCV\(_{2022}\) & 36.3  & 35.1  & 22.5  & 49.9  & 67.5  & 86.1  & 42.1  & 57.0 \\
          & & ACL  & ECCV\(_{2022}\) & 41.2  & 41.8  & 20.4  & 45.9  & \underline{74.3}  & \underline{89.3}  & 45.3  & 59.0 \\
          & & ReFID & TOMM\(_{2024}\) & 33.3 & 34.8 & 18.3  &39.8 & 67.6 & 85.3  & 39.7 & 53.3\\
          & & GMN & TCSVT\(_{2024}\) &  \underline{43.2} &  \underline{42.1} &  24.4 &  50.9 &  72.3 &  87.1 &  \underline{46.6} &  \underline{60.0} \\
        \cline{2-12}

            & \multirow{4}[0]{*}{ViT}  & ViT-B\(^*\) & ICLR\(_{2021}\)& 36.5 & 35.8 & 20.5 & 42.7 & 59.2 & 78.3 & 38.7 & 52.3\\

        & &  TransReID\(^*\)  & ICCV\(_{2021}\) & 36.5 & 36.1 & 23.2 & 46.3 & 59.9 & 79.8 &39.9  & 54.1  \\
            & &  CLIP-ReID\(^*\)  & AAAI\(_{2023}\) & 42.1 & 41.9 & \underline{26.6} & \underline{53.1} & 68.8 & 84.4 & 45.8 & 59.8\\
          & & \textbf{Ours} & This paper &  \textbf{44.4}    &   \textbf{44.6}   &    \textbf{31.1}  &   \textbf{59.4}   &   \textbf{79.4}    &  \textbf{91.3}    &  \textbf{51.6\(_{\scaleto{{51.5}}{3.0pt}} ^{\scaleto{{51.8}}{3.0pt}}\)}  & \textbf{65.1\(_{\scaleto{{65.0}}{3.0pt}} ^{\scaleto{{65.2}}{3.0pt}}\)} \\
       \hline
       \hline
        \multirow{13}[0]{*}{Protocol-3}  &\multirow{8}[0]{*}{CNN}& QAConv\(_{50}\) & ECCV\(_{2020}\) & 32.9  & 33.3  & 17.6  & 46.6  & 66.5  & 85.0    & 39.0    & 55.0 \\
          & & M\(^3\)L &CVPR\(_{2021}\)  & 35.7  & 36.5  & 17.4  & 38.6  & 62.4  & 82.7  & 38.5  & 52.6 \\
          & & MetaBIN &CVPR\(_{2021}\) & 43.0    & 43.1  & 18.8  & 41.2  & 67.2  & 84.5  & 43.0    & 56.3 \\
          & & META &ECCV\(_{2022}\) & 47.1  & 46.2  & 24.4  & 52.1  & 76.5  & 90.5  & 49.3  & 62.9 \\
          & & ACL  &ECCV\(_{2022}\) & 49.4  & \textbf{50.1}  & 21.7  & 47.3  & 76.8  & \underline{90.6} & 49.3  & 62.7 \\
          & & META+IL  &TMM\(_{2023}\) & 48.9 & \underline{48.8} & \underline{26.9}  & \underline{54.8} & \underline{78.9} &  \textbf{91.2} &  \underline{51.6} & \underline{64.9} \\
          & & ReFID & TOMM\(_{2024}\) & 45.5 &  44.2 &  20.6 &  43.3 &  72.5 &  87.9 & 46.2 & 58.5\\
          & & GMN & TCSVT\(_{2024}\) &  \underline{49.5} &  \textbf{50.1} &  24.8 &  51.0 &  75.9 &  89.0 &  50.1 &  63.4\\
       \cline{2-12}
            & \multirow{4}[0]{*}{ViT}  & ViT-B\(^*\) & ICLR\(_{2021}\)& 39.4 &39.4 & 20.9 & 43.1 & 63.4 & 81.6 & 41.2 & 54.7\\
            & &  TransReID\(^*\)  & ICCV\(_{2021}\) & 44.0 & 45.2 & 23.4 & 46.9 & 63.6 & 82.5 & 43.7 &  58.2 \\
            & &  CLIP-ReID\(^*\)  & AAAI\(_{2023}\) & 44.9 & 45.8 & 26.8 & 52.6 & 67.5 & 83.4 & 46.4 &60.6\\
          & & \textbf{Ours} & This paper& \textbf{50.1}   & \textbf{50.1}   &    \textbf{32.9}   &   \textbf{61.4}   & \textbf{79.8}    & \textbf{91.2}   &  \textbf{54.3\(_{\scaleto{{53.5}}{3.0pt}} ^{\scaleto{{55.0}}{3.0pt}}\)}  & \textbf{67.6\(_{\scaleto{{66.9}}{3.0pt}} ^{\scaleto{{68.3}}{3.0pt}}\)} \\
       \Xhline{1px}
    \end{tabular}%

\label{tab:2}

\end{table*}%

\textbf{Implementation details.}
All experiments are conducted using the same settings.
Most experiments are run on an RTX 3090 24GB GPU.
Some experiments are performed on a RTX 4090 24GB GPU.
But we set up the same environment on different GPUs, including Python version, Pytorch version, NumPy version, and so on.
To be specific, we employ a batch-size of 128 throughout our experiments.
Training took place in three stages: the first stage lasted for 3 epochs, the second stage for 150 epochs (120 for learning ID-specific tokens and 30 for learning Domain-specific tokens), and the third stage for an additional 60 epochs.
We will provide more details about other experimental hyperparameters in the following sections.

\subsection{Comparison with State-of-the-art Methods}
We compare our method with state-of-the-art (SOTA) methods in generalizable person re-identification, including QAConv\(_{50}\)~\cite{Liao_Shao_2020}, M\(^{3}\)L~\cite{Zhao_Zhong_Yang_Luo_Lin_Li_Sebe_2021}, MetaBIN~\cite{Choi_Kim_Jeong_Park_Kim_2021}, META~\cite{xu2022mimic}, ACL~\cite{Zhang_Dou_Yunlong_Li}, IL~\cite{Tan_Wang_Ding_Gong_Jia_2022}, ReFID~\cite{peng2024refid} and GMN~\cite{qi2024generalizable}.
Additionally, we also compare our model with models based on Vision Transformers (ViT) as the backbone, specifically the ViT-B model, which has approximately 86 million parameters.
Including, ViT~\cite{ViT}, TransReID~\cite{He_Luo_Wang_Wang_Li_Jiang_2021}(Since we are dealing with cross-domain person re-identification, where the camera IDs in the source and target domains do not match, the SIE component of the TransReID framework is not used.), CLIP-ReID~\cite{li2023clip}, PAT~\cite{ni2023part} and DSM+SHS~\cite{Li2023StyleControllableGP}.

As shown in Tab.~\ref{tab:1} and Tab.~\ref{tab:2}, our method consistently outperformed other methods in all three protocols.
The results for our method in the table represent the averages of three experiments conducted under the same settings.
Compared to other methods, our method demonstrated superior performance.
And these demonstrate that our method performs effectively in generalizable person re-identification tasks.

\textbf{Protocol-1.}
Comparing our method to other SOTA methods in the context of Protocol-1, it becomes evident that our model exhibits superior performance.
\textbf{ACL}, which is one of the top-performing methods in this comparison, achieved an mAP of 73.4\% and an R1 of 63.0\%, while DSM+SHS, another competitive approach, achieved an mAP of 74.1\% and an R-1 of 66.0\%.
In both mAP and R1, our method outperforms these models with an mAP of \textbf{82.2\%} and an R1 of \textbf{75.5\%}.
Moreover, compared to the SOTA result \textbf{DSM+SHS}, which is based on ViT-B, our method demonstrates similarly remarkable superiority and mAP increased by \textbf{8.1\%}.

\textbf{Protocol-2 \& Protocol-3.}
In the context of Protocol-2 and Protocol-3.
Our method outperforms other SOTA methods in terms of average mAP and average R1.
Especially under Protocol-2, our method exhibits results that are unmatched by other methods.
It achieved an impressive average mAP of \textbf{51.6\%} and a remarkable R1 of \textbf{65.1\%}, demonstrating its remarkable effectively.
In comparison, the prior state-of-the-art models, such as \textbf{GMN} with an average mAP of 46.6\% and an R1 of 60.0\%, showed notable but comparatively lower performance.
Besides, the experiments under Protocol-3, while not showcasing the comprehensive superiority observed in Protocol-2, our method still achieves the highest evaluations in terms of average mAP and average R1.
In Protocol-3, the results highlight our method's effectiveness in various scenarios, it achieved an impressive average mAP of \textbf{54.3\%} and a remarkable R1 of \textbf{67.6\%}.

\begin{table*}[ht]

  \centering
  \caption{Studies of epochs. To compare with the baseline, we denote the three stages as ``Initial Stage'', ``Stage-1'' and ``Stage-2''. The ``Stage-1'' encompasses the process of learning ID-specific tokens and learning Domain-specific tokens, represented as ``I'' and ``D'' respectively. Due to the absence of initial stage and the process of learning domain-specific tokens in the baseline, they are represented by ``-''.}

  \begin{tabular}{c|c|cc|c|cccc|c}
\Xhline{1px}
    \multirow{2}[0]{*}{Method} & \multirow{2}[-1]{*}{Initial} &\multicolumn{2}{c|}{Stage-1 epochs}  & \multirow{2}[-1]{*}{Stage-2} & \multicolumn{4}{c|}{Avergae} & \multirow{2}[0]{*}{Train time (under 3090 gpu)}  \\
\cline{3-4} \cline{6-9}     &  ~~~epochs~~~   &  ~~~ID~~ & ~~DM~~ &   ~~~epochs~~~   & ~~~mAP~~~   &~~~R1~~~&~~~R5~~~&~~~R10~~~ & \\

    \hline
    \hline
~~~~baseline~~~~       & -     & 120   & -     & 60    & 69.8   &  60.2  &   82.2   & 87.5   & 467 (min) \\
\hdashline       \multirow{2}[0]{*}{\text{ours}}   & 3     & 96    & 24    & 57    &   80.8 &  	73.6  & 87.9 &   92.1  & 473 (min) \\
       & 10    & 96    & 24    & 50    &   80.8 &  	72.6  & 86.6 & 91.8 & 470 (min) \\
\Xhline{1px}
    \end{tabular}%
  \label{tab:4}%

\end{table*}%

\begin{table}[t]
  \centering
  \caption{Ablation studies of our method. Where ``baseline'' represents CLIP-ReID; ``A'' represents the use of \textbf{three-stage strategy}; ``B'' represents using the \textbf{bidirectional guiding method}; ``C'' represents the using of \(\mathcal{L}_{apn}\).}
  \begin{tabular}{cc|cc }
\Xhline{1px}
    \multirow{2}[0]{*}{Setting} & \multirow{2}[0]{*}{Method} & \multicolumn{2}{c}{Avergae} \\
 \cline{3-4}          &       & ~~mAP~~ & ~~~R1~~~ \\
\hline
\hline
    \multirow{5}[0]{*}{Protocol-1} & Baseline & 69.8 & 60.2 \\
         &    Baseline+A   & 76.9 & 67.4 \\
          &    Baseline+B   &75.8  & 67.2 \\
          &    ~~~Baseline+A+B \text{w/o} C ~~~ & 80.7 & 73.4 \\
         &    Baseline+A+B  & \textbf{82.2} & \textbf{75.5} \\
\hline
    \multirow{5}[0]{*}{Protocol-2} &    Baseline   &  45.8     & 59.8 \\
        &   Baseline+A    &    50.6   & 64.4 \\
         &   Baseline+B     &   48.1  & 62.2  \\
        &    Baseline+A+B \text{w/o} C  & 51.4 & 64.8 \\
       &    Baseline+A+B &  \textbf{51.7}  & \textbf{65.1} \\
\Xhline{1px}
    \end{tabular}%
  \label{tab:3}
    \vspace {-1em}
\end{table}

\subsection{Ablation Studies}

\textbf{Training time analysis.}
Admittedly, compared to baseline, our method does involve an increase in the total training duration due to the additional initial training stage and the training of domain-specific tokens. 
However, to further validate the superiority of our method, we plan to balance the epochs, conducting experiments without increasing the overall number of epochs.
We experiment with evenly distributing the training epochs as shown in the Tab.~\ref{tab:4}.
In other words, the number of epochs added in the first stage is balanced by reducing the epochs in the third stage.
Similarly, the additional epochs for learning domain-specific tokens in the second stage are compensated for by a reduction in the epochs dedicated to learning ID-specific tokens.
This approach ensured that the total training epochs remained unchanged.
According to the experimental results in Tab.~\ref{tab:4}, compared to the baseline, our method can achieve significant improvements without increasing the training epochs (time).

\textbf{Effectiveness of components.}
To further investigate the impact of our method, we conduct a series of ablation experiments, focusing on both Protocol-1 and Protocol-2.
These ablation experiments are specifically designed to comprehensively evaluate the effectiveness of our method.
Since Protocol-2 and Protocol-3 only differ in terms of data volume, we opted not to conduct additional experiments on Protocol-3.

From the results presented in Tab.~\ref{tab:3}, the effectiveness of each component of our method can be observed.
On one hand, the results of ``Baseline+A'' indicate that our proposed \textbf{three-stage learning method} is indeed effective in more accurately learning from text prompts, successfully addressing the first problem raised in introduction.
On the other hand, the results of ``Baseline+B'' signify the effectiveness of the \textbf{bidirectional guiding method} in prompting the image encoder to learn domain-invariant features, addressing the second problem posed in our introduction.
The comprehensive evaluation of ``Baseline+A+B'', demonstrates promising results, affirming the effectiveness of our method.
These demonstrate that all components of our proposed method are effective for cross-domain person re-identification tasks.

\subsection{Further Analysis}

\textbf{Performance on BLIP.}
We attempt to apply our proposed three-stage learning strategy and bidirectional guiding method to other VLMs.
We choose BLIP for application.
As shown in Tab.~\ref{tab:6}, we try to replace CLIP in the CLIP-ReID framework with BLIP (we name it BLIP-ReID) and obtain results of mAP 72.8\% and rank-1 62.8\% under the setting of protocol-1.
Subsequently, we apply our method to BLIP-ReID (named BLIP-FGDI), which leads to a 2.5\% improvement in mAP, reaching 75.3\%, and a 3.0\% improvement in rank-1, reaching 65.8\%.
It can be seen that our method has the potential to enhance the generalization performance on other VLMs.

\begin{table}[t]

    \begin{minipage}[t]{0.42\linewidth}
                \caption{Experiments on application of our method to BLIP.}
    \setlength\tabcolsep{4pt}
        \centering
                \begin{tabular}{l cc}
        \Xhline{1px}

     \multirow{2}[0]{*}{Method}& \multicolumn{2}{c}{Average}\\
                                & mAP &R1\\

        \hline
        BLIP &72.8 & 62.8 \\
        BLIP-FGDI & 75.3\(^{\uparrow{2.5}}\) & 65.8\(^{\uparrow{3.0}}\) \\
    \Xhline{1px}

    \end{tabular}
    \label{tab:6}

    \end{minipage}
    \hfill
    \begin{minipage}[t]{0.5\linewidth}
         \caption{Fine-tuning ViT model with CLIP's pretrained weights.}

    \setlength\tabcolsep{4pt}
        \centering
         \begin{tabular}{l cc}
        \Xhline{1px}

     \multirow{2}[0]{*}{Method}& \multicolumn{2}{c}{Average}\\
                                & mAP &R1\\
        \hline
        ViT-B w/ C &   70.5    &  60.5 \\
        CLIP-FGDI & 81.6\(^{\uparrow{11.1}}\) & 74.3\(^{\uparrow{13.8}}\) \\
    \Xhline{1px}

    \end{tabular}
  \label{tab:7}

    \end{minipage}

\end{table}

\textbf{Fine-tuning ViT model with CLIP's pretrained weights.}
We have to acknowledge that leveraging the powerful performance of CLIP yielded promising results.
However, the ablation experiments demonstrate the effectiveness of our method, showing a significant improvement of 11.8\% compared to the baseline (CLIP-ReID).
Moreover, to provide a comparison, we include an experiment featuring a ViT model utilizing CLIP's pretrained weights, as shown in Tab.~\ref{tab:7}.
It needs to be note that ``ViT-B w/ C'' denotes fine-tuning with CLIP's image encoder pretraining weights while the backbone network is ViT-B.

\begin{table}[ht]
  \centering
  \caption{Improvements of our method on different backbone as image encoder under Protocol-1.}

      \setlength\tabcolsep{8pt}
    \begin{tabular}{l|cccc}
        \Xhline{1px}
    \multicolumn{1}{c|}{\multirow{2}[0]{*}{Method}}& \multicolumn{4}{c}{Avergae} \\
        \cline{2-5}
        &   ~~~mAP~~~  & ~~R1~~ & ~~R5~~   & ~~R10~~ \\
        \hline
        \hline
        ResNet-50~~~~ &   45.8   &   35.3   &  57.2    & 67.1  \\
        +ours &      \textbf{52.3\(^{\uparrow{6.5}}\)}       & \textbf{41.0\(^{\uparrow{5.7}}\)}     &  \textbf{65.0\(^{\uparrow{7.8}}\)}    & \textbf{73.5\(^{\uparrow{6.4}}\)}   \\
            \hline
        ResNet-101 &   57.0   &  44.2    &   71.9   & 82.6 \\
        +ours &      \textbf{59.9\(^{\uparrow{2.9}}\)}       & \textbf{46.9\(^{\uparrow{2.7}}\)}     &  \textbf{76.6\(^{\uparrow{4.7}}\)}    & \textbf{84.0\(^{\uparrow{1.4}}\)}   \\
                \hline
        ViT-B-16 &      69.8  &  60.2   &   82.2   & 87.5  \\
        +ours &     \textbf{81.6\(^{\uparrow{11.8}}\)}       & \textbf{74.3\(^{\uparrow{14.1}}\)}     &  \textbf{89.7\(^{\uparrow{7.5}}\)}    & \textbf{94.1\(^{\uparrow{6.6}}\)}   \\
                \hline
        ViT-B-32 &     65.8  &  55.4    &   78.8   & 83.6  \\
        +ours &       \textbf{72.8\(^{\uparrow{7.0}}\)}       & \textbf{63.3\(^{\uparrow{7.9}}\)}     &  \textbf{85.3\(^{\uparrow{6.5}}\)}    & \textbf{88.8\(^{\uparrow{5.2}}\)}   \\
    \Xhline{1px}
    \end{tabular}%
  \label{tab:8}%
\end{table}%

\textbf{Different Backbone.}
To further validate the effectiveness of our method, we conduct experiments using various backbone as image encoder, including ResNet-50, ResNet-101, ViT-B-16 and ViT-B-32.
These experiments aimed to assess the effectiveness of our method across a spectrum of backbone models.
Upon analyzing the results presented in Tab.~\ref{tab:8}, it becomes evident that our method consistently demonstrates a significant improvement in performance, irrespective of the specific backbone architecture employed.

\textbf{Further Analysis into the BGM.}
The Bidirectional Guiding Method (BGM) is an approach that leverages both domain-invariant and domain-relevant features to jointly guide the model in learning domain-invariant image features.
This is specifically manifested through the \emph{apn loss}.
In this context, the image encoder's output of image features serves as the \emph{anchor}, where a domain-invariant prompt generated by the Text encoder is utilized as the \emph{positive} example.
Meanwhile, a domain-variant prompt generated by the Text encoder is used as the \emph{negative} example.
This process involves calculating the triplet loss to bidirectionally guide the Image encoder in its ability to extract domain-invariant features.

The crucial aspect lies in how to compute the measure between the \emph{anchor} and both the \emph{positive} and \emph{negative}, specifically denoted as ``\(\text{Sim}\)'' in Eq~\eqref{eq8}.
We employ three methods to implement the \emph{apn loss}, which we named as \emph{euclidean distance based apn loss} (ED-based), \emph{cosine similarity based apn loss} (CS-based) and \emph{contrastive based apn loss} (Con-based).
Moreover, we also attempted to combine the \(\mathcal{L}{apn}\) and \(\mathcal{L}{i2tce}\) into a single loss function, namely \(\mathcal{L}{apnce}\).
Additionally, we present the corresponding parameters and experimental results in Tab.~\ref{tab:8}.

\begin{figure}[b]
  \centering
   \includegraphics[width=0.7\linewidth]{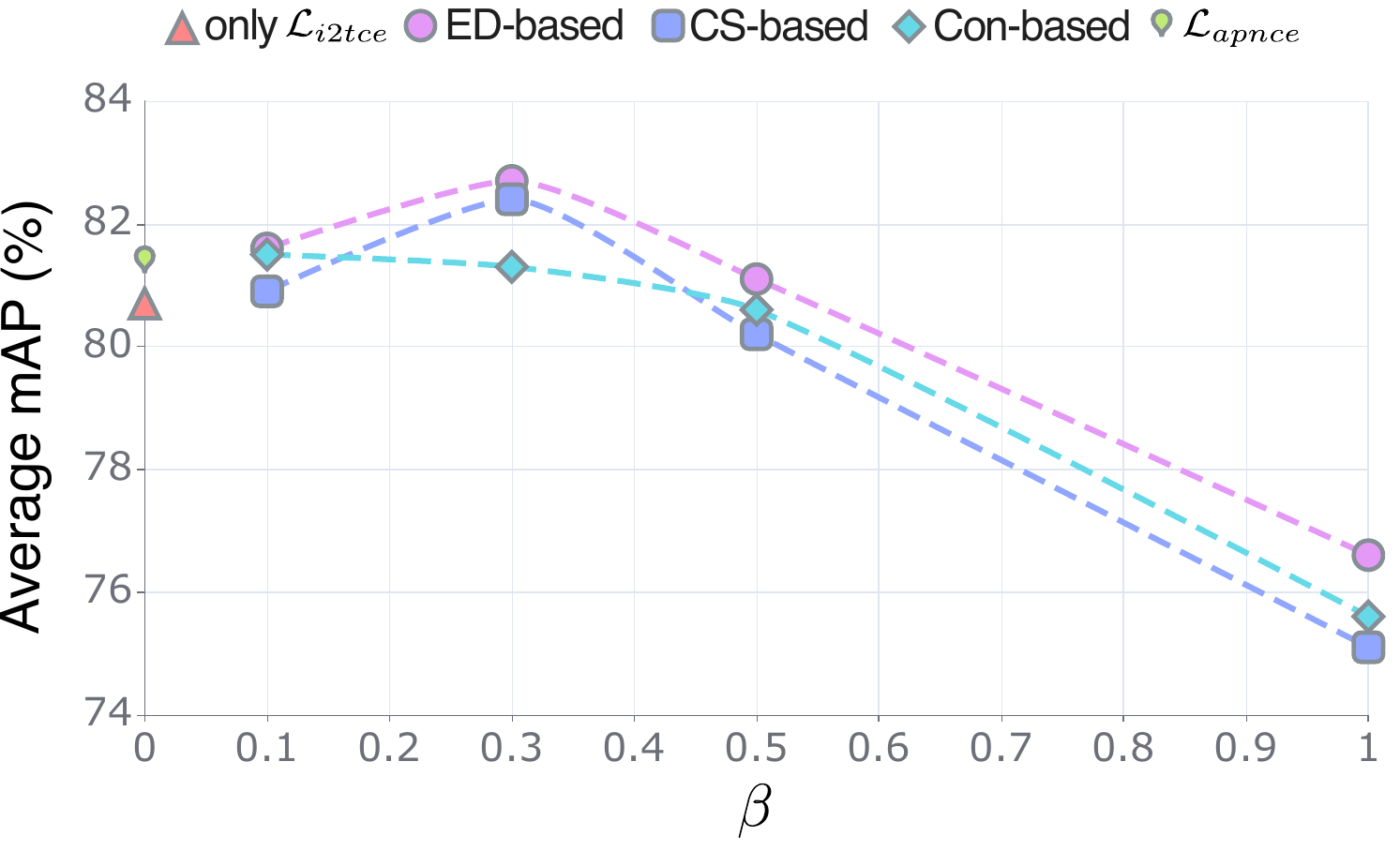}
   \caption{Line charts illustrating the results for different loss functions and varying values of the hyper-parameter \(\beta\).}
  \label{fig:3}

\end{figure}
\begin{figure}[b]
      \centering
    \includegraphics[width=0.7\linewidth]{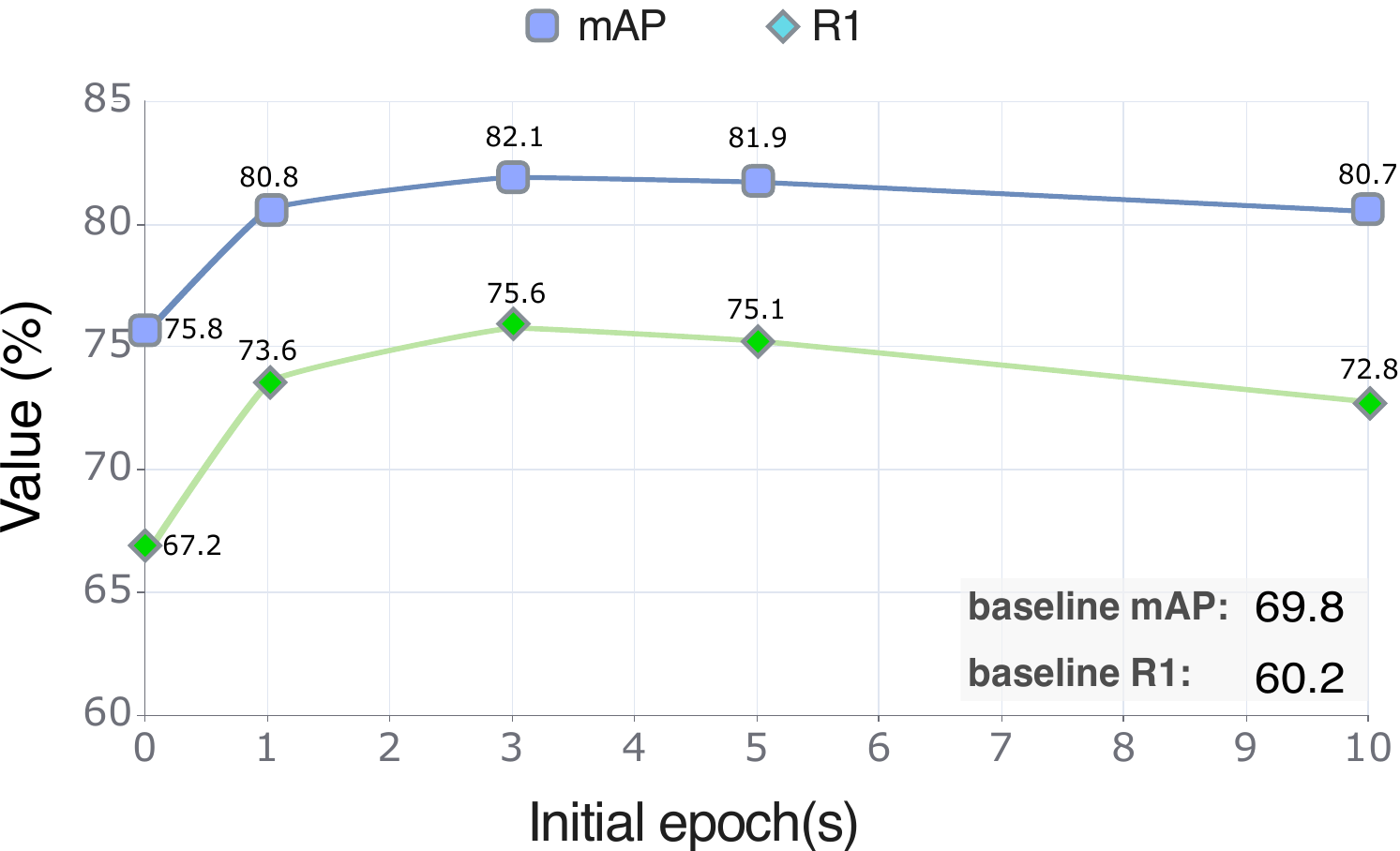}
    \caption{Line graph of experimental results for different initialization epochs.}
    \label{fig:4}
\end{figure}

Euclidean distance is a simple and versatile method for measuring the straight-line distance between two points.
It is widely used in various fields due to its simplicity and effectiveness in preserving geometric relationships.
Therefore, we attempt to implement a distance metric using Euclidean distance for \(F^a \in \mathbb{R}^{B \times d}\) (anchor feature), \(F^{p} \in \mathbb{R}^{B \times d}\) (positive feature), and \(F^n \in \mathbb{R}^{B \times d}\) (negative feature), namely:
\begin{eqnarray}
    \text{Sim\(_{\text{ED}}\)}(x, y) = \sqrt{\sum_{i=1}^{d} (x_i - y_i)^2},
\end{eqnarray}
aims to bring the \(F_i^a\) and \(F_i^{p}\) closer and simultaneously push the \(F_i^a\) and \(F_i^n \) further apart.

In the second stage of training for learning the prompt, we utilize \(\mathcal{L}{i2t}\) and \(\mathcal{L}{t2i}\) to unify the text feature space and the image feature space, as shown in Eq.~\eqref{eq2}.
In Eq~\eqref{eq2}, ``s'' represents an operation that measures the similarity between image and text features.
Specifically, it is implemented using the \emph{dot product} operation, which is particularly useful for assessing the directional similarity or correlation between vectors.
In contrast, the euclidean distance reflects the actual distance in space, which is related to the absolute size of vectors.
On the other hand, cosine similarity partially measures the similarity between two vectors in terms of their directions:
\begin{eqnarray}
    \text{Sim\(_{\text{CS}}\)}(x, y) = \frac{x \cdot y}{\|x\| \cdot \|y\|}.
\end{eqnarray}
If the directions of two vectors are close, both their dot product and cosine similarity will be relatively large.

In addition to implementing the apn loss using a triplet-based approach, another solution involves designing loss based on a contrastive method.
For inputs \(F^a \in \mathbb{R}^{B \times d}\), \(F^{p} \in \mathbb{R}^{B \times d}\), and \(F^{n*} \in \mathbb{R}^{M \times d}\), we first combine \(F^{p}\) and \(F^{n*}\) into \(F^{pn} \in \mathbb{R}^{(B+M) \times d}\).
Subsequently, we calculate the corresponding contrastive loss:
\begin{eqnarray}
    \mathcal{L}_{apn}(F^a,F^{pn}) = - \frac{1}{B} \sum_{i=1}^B \log\frac{e^{s(F^a_i, F^{pn}_i)}}{\sum_{k=1}^{B+M} e^{s(F^a_i, F^{pn}_k)}} ,
\label{eq13}
\end{eqnarray}
where \( s(\cdot) \) represents the dot product operation, \(B\) denotes the batch size, \(M\) represents the number of unique IDs in all samples, \(d\) is the feature dimension, and \(F^{n*}\) denotes all domain-relevant prompt features.

\begin{table*}[ht]
  \centering
  \caption{Further analysis into the BGM.}
  \setlength\tabcolsep{9pt}
    \begin{tabular}{c|c|c|cccccccccc}
\Xhline{1px}
    \multirow{2}[0]{*}{\(\mathcal{L}_{apn}\)} & \multirow{2}[0]{*}{~~\(\mathcal{L}_{i2tce}\)~~} & \multirow{2}[0]{*}{~~~~\(\beta\)~~~~} & \multicolumn{2}{c}{PRID} & \multicolumn{2}{c}{GRID} & \multicolumn{2}{c}{VIPeR} & \multicolumn{2}{c}{iLIDs} & \multicolumn{2}{c}{Average} \\
    &       &       & \multicolumn{1}{c}{~mAP~} & \multicolumn{1}{c}{~R1~} & \multicolumn{1}{c}{~mAP~} & \multicolumn{1}{c}{~R1~} & \multicolumn{1}{c}{~mAP~} & \multicolumn{1}{c}{~R1~} & \multicolumn{1}{c}{~mAP~} & \multicolumn{1}{c}{~R1~} & \multicolumn{1}{c}{~mAP~} & \multicolumn{1}{c}{~R1~} \\
    \hline
    \hline
    -     &   \checkmark    & 0     & 77.2   &    	70.0   &    	75.1   &    	66.4	   &    80.1   &    	72.8   &    	90.2   &    	84.3     &      80.7 & 73.4 \\
    \hline
    \hline
    \multirow{4}[0]{*}{~~ED-based~~} & \multirow{3}[0]{*}{\checkmark} & 0.1   &    77.7   &   68.0    &   75.2    &    67.2   &   81.7    &    75.6   &    91.9   &   88.3    &    81.6   &  74.8 \\
\cline{3-3}          &       & 0.3   &   80.2   &    71.0   &    75.8   &    68.0   &   83.0    &   77.5    &    91.7   &   88.3    &   82.7    & 76.2 \\
\cline{3-3}          &       & 0.5   &    78.2   &    68.0   &   74.4    &   65.6    &   81.1    &   74.1    &    90.6   &   86.7    &   81.1    & 73.6 \\
\cline{2-3}          & -     & 1.0   &    68.8   &    56.0   &   70.8   &    60.0   &   80.7    &    72.5   &    86.0   &      80.0  &    76.6   & 67.1 \\
    \hline
    \hline
    \multirow{4}[0]{*}{CS-based} & \multirow{3}[0]{*}{\checkmark} & 0.1   &   77.6    &    68.0   &   75.9    &   69.6    &  81.4    &    75.3   &   88.5    &   81.7    &    80.9   &  73.6 \\
\cline{3-3}          &       & 0.3   &   82.4    &   74.0    &   74.2    &   64.8    &   81.1    &    75.0   &     91.9  &   88.3    &    82.4   &  75.5 \\
\cline{3-3}          &       & 0.5   &    77.3   &    66.0   &   72.4    &    62.4   &   81.8    &   75.6    &    89.1   &    83.3   &    80.2   & 71.8  \\
\cline{2-3}          & -     & 1.0   &     68.3  &    56.0   &    67.3   &     55.2  &    78.1   &    69.9   &   86.9    &    81.7   &  75.1     & 65.7 \\
    \hline
    \hline
    \multirow{4}[0]{*}{Con-based} & \multirow{3}[0]{*}{\checkmark} & 0.1   &   81.6    &  74.0    &   74.8    &   64.8    &    80.7   &   74.1    &   88.9    &    83.3   &   81.5    &  74.0 \\
\cline{3-3}          &       & 0.3   &   77.9    &    69.0   &   77.0    &   69.6    &   82.9    &    76.6   &    87.5   &    81.7   &    81.3   & 74.2 \\
\cline{3-3}          &       & 0.5   &    77.7   &    68.0   &    74.5   &   65.6    &   81.9    &    75.3   &    88.5   &    81.7   &    80.6   & 72.6 \\
\cline{2-3}          & -     & 1.0   &   66.8    &   58.0    &   70.2    &    58.4   &     78.6  &    69.9   &    86.7   &    80.0   &    75.6   & 66.6 \\
\hline
\hline
    \multicolumn{3}{c|}{\(\mathcal{L}_{apnce}\)}   &   79.0    &    70.0   &   73.7    &    64.8   &   81.9    &   74.7   &   90.2    &    85.0   &   81.2    & 73.6 \\
\Xhline{1px}
    \end{tabular}%
  \label{tab:9}%
\end{table*}

\begin{figure*}[ht]

    \centering
    \begin{subfigure}{0.29\linewidth}
        \centering
        \includegraphics[width=\linewidth]{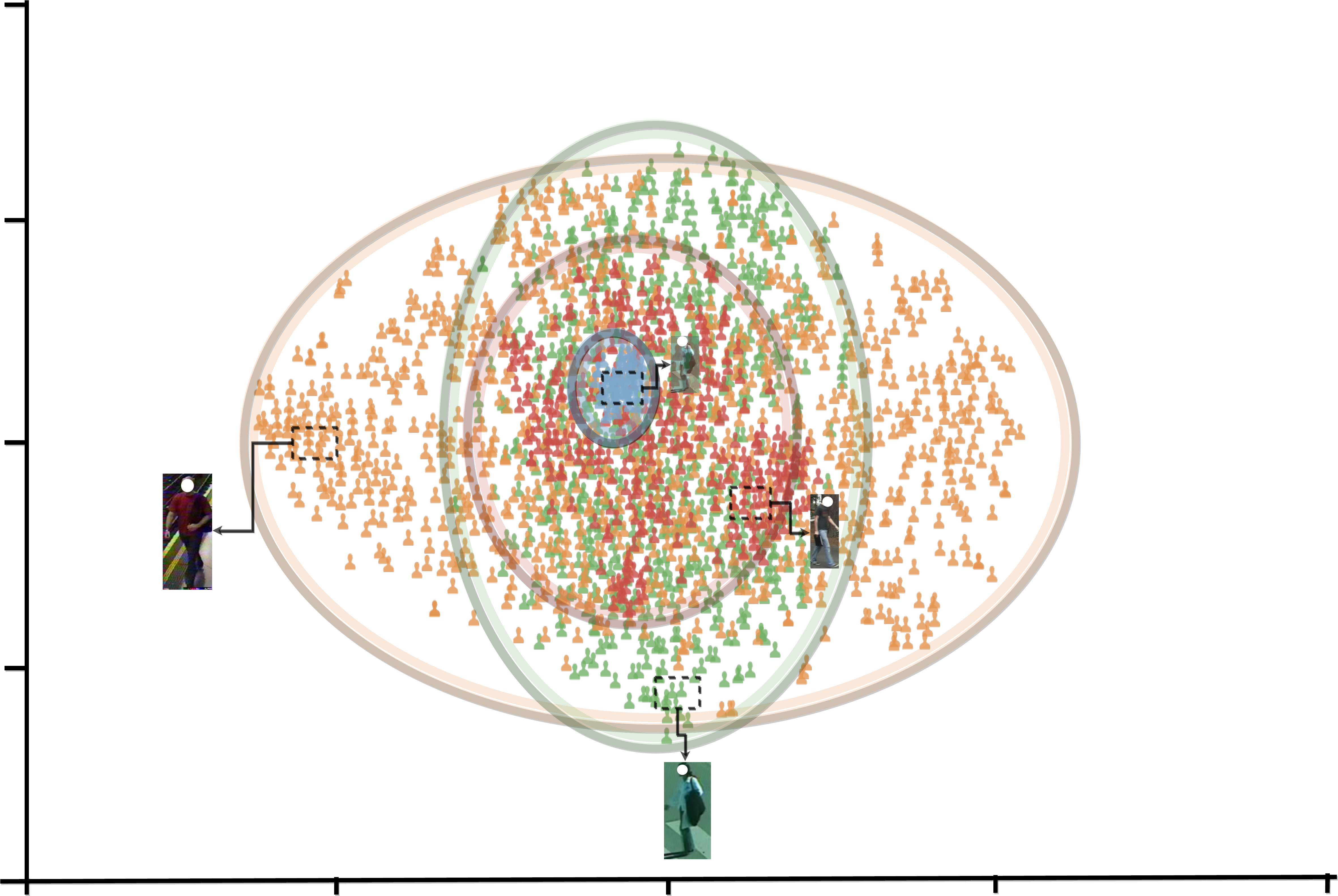}
        \caption{Baseline}
        \label{fig:5a}
    \end{subfigure}
    \hfill
    \begin{subfigure}{0.29\linewidth}
        \centering
        \includegraphics[width=\linewidth]{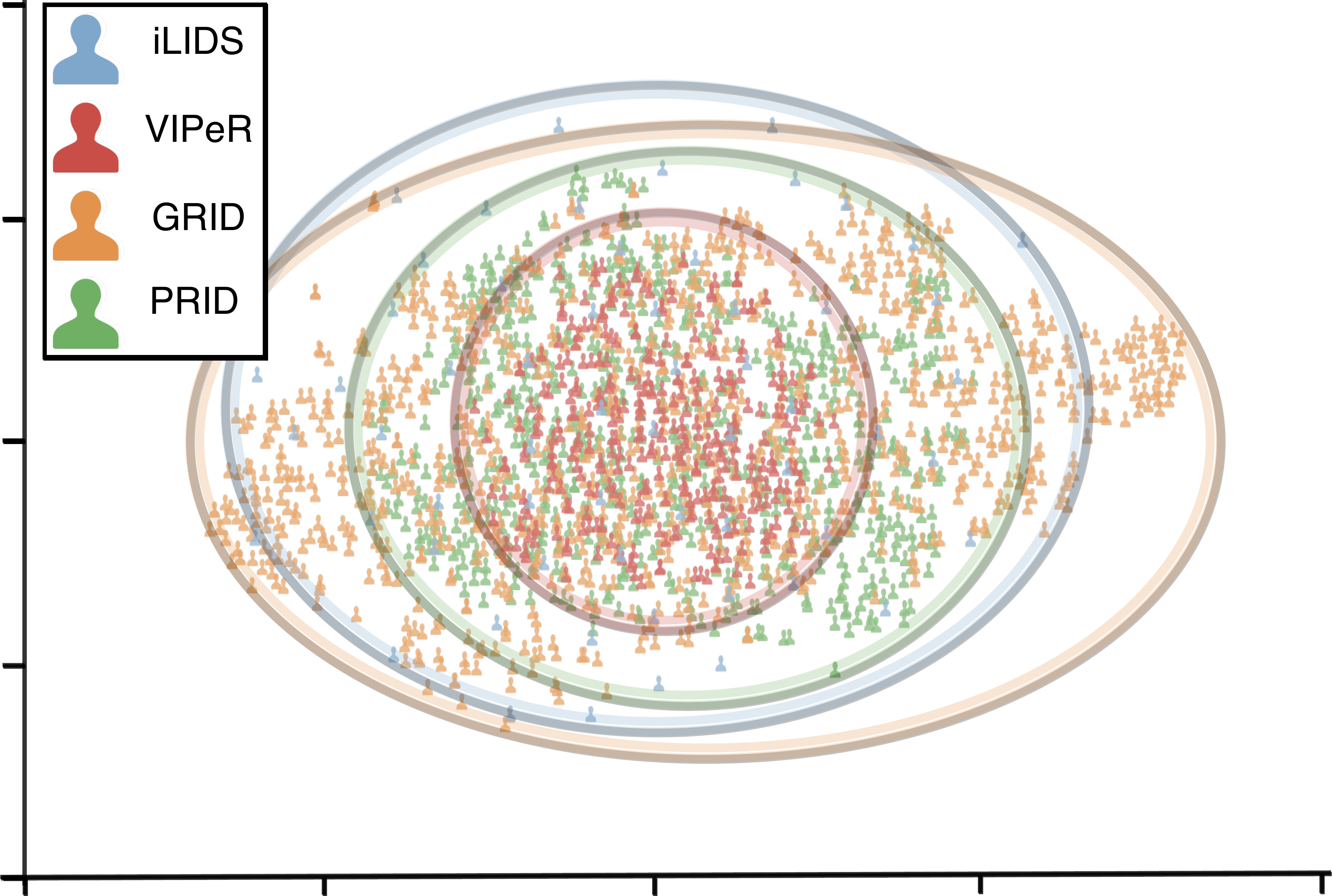}
        \caption{CILP-FGDI\(_{(\text{ours})}\)}
        \label{fig:5b}
    \end{subfigure}
    \hfill
    \begin{subfigure}{0.4\linewidth}
        \begin{subfigure}[b]{0.49\linewidth}
            \centering
            \includegraphics[width=\linewidth]{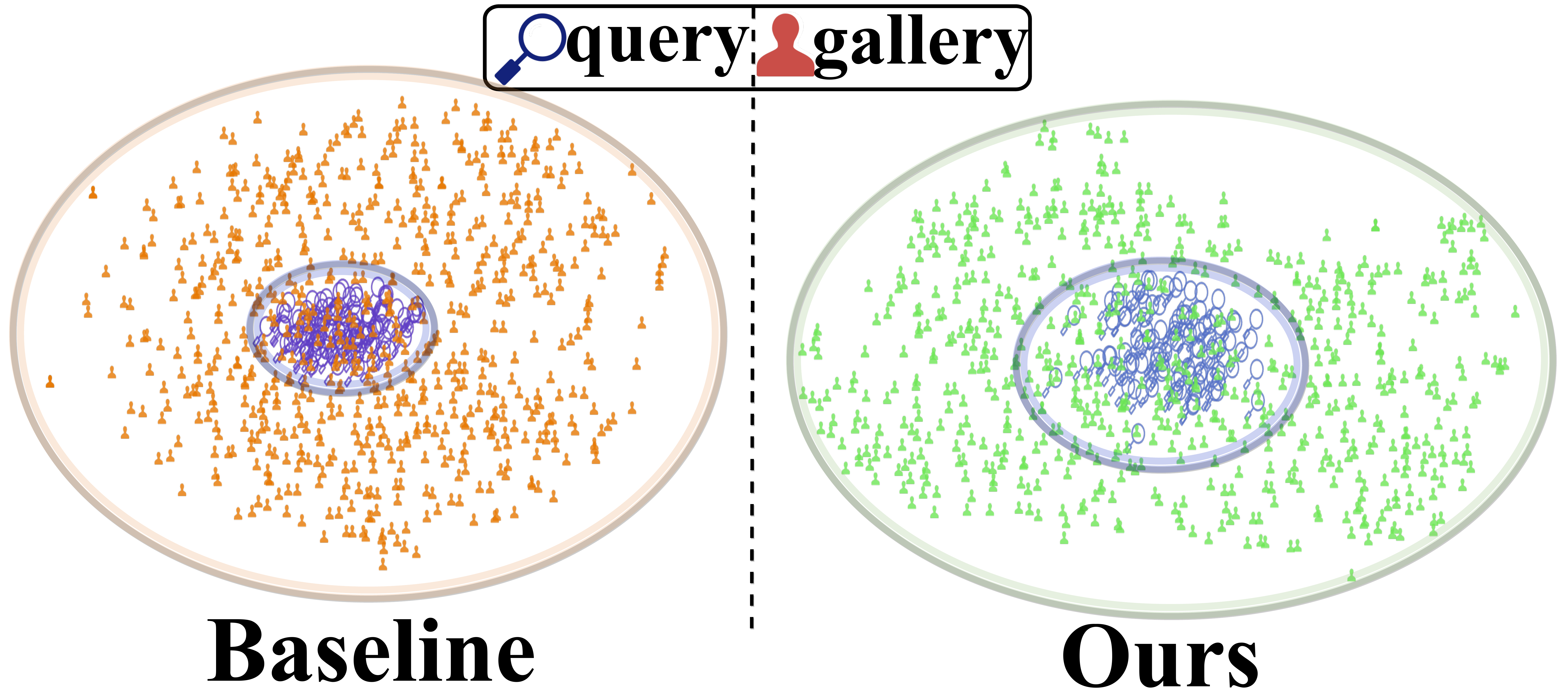}
            \caption{PRID}
            \label{fig:5c}
        \end{subfigure}
        \hfill
        \begin{subfigure}[b]{0.49\linewidth}
            \centering
            \includegraphics[width=\linewidth]{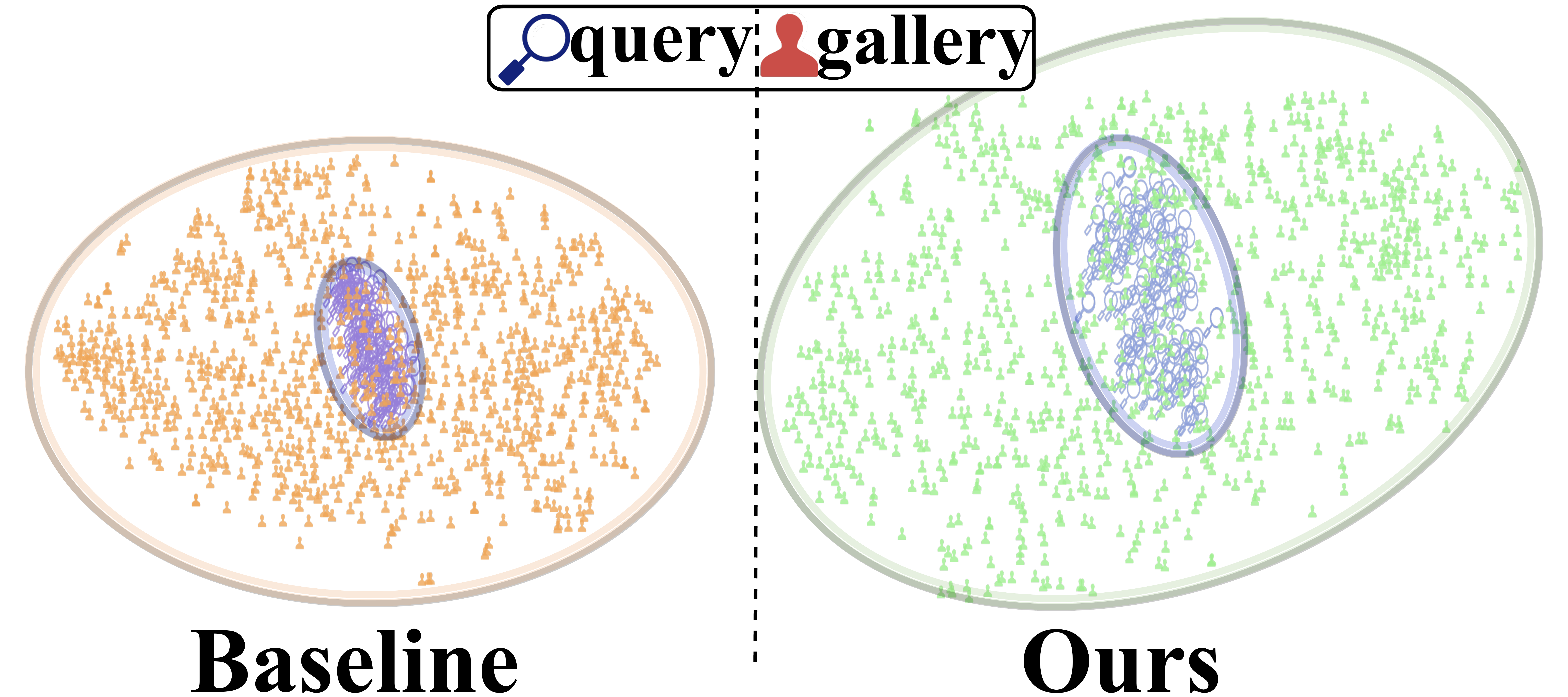}
            \caption{GRID}
            \label{fig:5d}
        \end{subfigure}

        \begin{subfigure}[b]{0.49\linewidth}
            \centering
            \includegraphics[width=\linewidth]{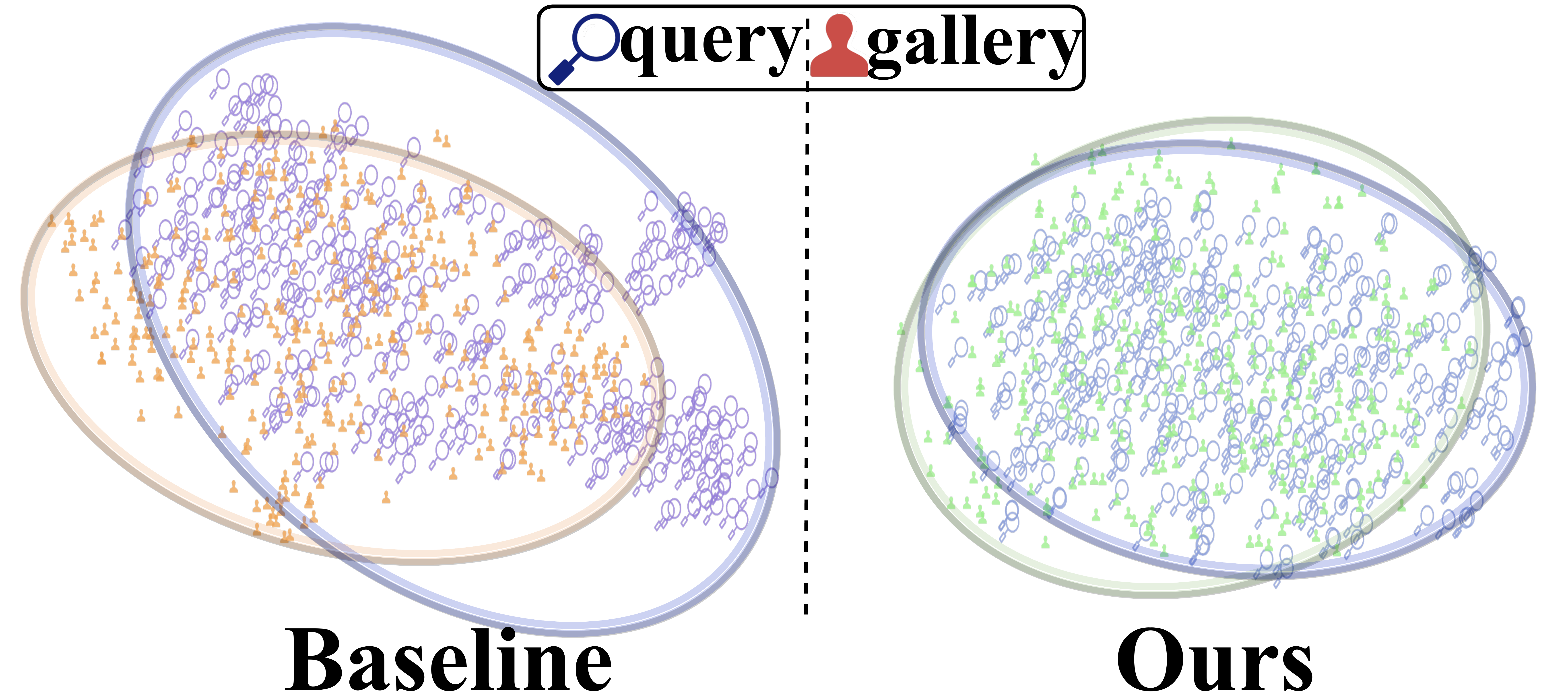}
            \caption{VIPeR}
            \label{fig:5e}
        \end{subfigure}
        \hfill
        \begin{subfigure}[b]{0.49\linewidth}
            \centering
            \includegraphics[width=\linewidth]{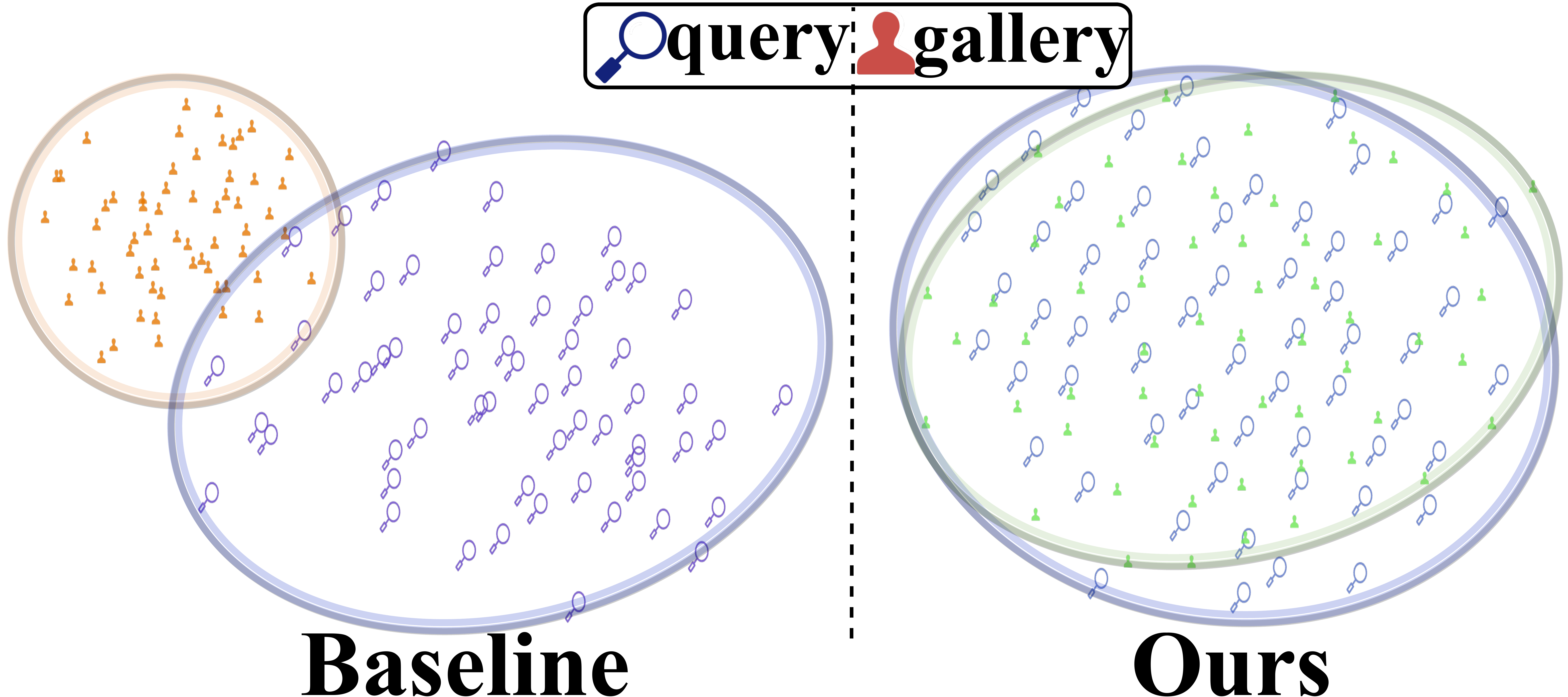}
            \caption{iLIDs}
            \label{fig:5f}
        \end{subfigure}
    \end{subfigure}

    \caption{The t-SNE visualization of embeddings on the target domains: (a) t-SNE results obtained using baseline; (b) t-SNE results obtained using our method. (c), (d), (e) and (f) are scatter plots comparing the baselines with our method on PRID, GRID, VIPeR and iLIDs datasets, respectively. Best viewed in colors.}
    \label{fig:5}

\end{figure*}

Moreover, by observing Eq.~\eqref{eq13} and Eq.~\eqref{eq4}, we attempt to integrate the objectives of Eq.~\eqref{eq13} with Eq.~\eqref{eq4}, seeking the combination of both \(\mathcal{L}_{apn}\) and \(\mathcal{L}_{i2tce}\) loss functions.
For inputs \(F^a \in \mathbb{R}^{B \times d}\), \(F^{p*} \in \mathbb{R}^{M \times d}\), and \(F^{n*} \in \mathbb{R}^{M \times d}\).
And the combination of \(F^{p*}\) and \(F^{n*}\) forms \(F^{pn*} \in \mathbb{R}^{2M \times d}\).
We name the new loss function \(\mathcal{L}_{apnce}\), as fellows:
\begin{eqnarray}
\mathcal{L}_{apnce}(i) = -\sum_{k=1}^N q_k\log\frac{e^{s(F^a_i, F^{p*}_{k})}}{\sum_{j=1}^{2M} e^{s(F^a_i, F^{pn*}_{j})}} .
\end{eqnarray}

\textbf{Hyperparameter analysis.}
Under the experimental settings of Protocol-1, we conduct a comprehensive experimental study on BGM, particularly focusing on various parameter and loss configurations, addressing the scenarios mentioned above.
The experiments related to hyper-parameters primarily emphasized the analysis of \(\beta\) in Eq.~\eqref{eq9}.
Through Tab.~\ref{tab:8} and Fig.~\ref{fig:3}, it can be observed that the three forms of \(\mathcal{L}{apn}\) yield satisfactory results when the hyperparameter \(\beta\) is around 0.3.
Specifically, when using the \(\mathcal{L}{apn}\) based on Euclidean Distance, the highest results are achieved.
The average mAP reaches 82.7\%, and the average R1 reaches 76.2\%.
Simultaneously, the Cosine Similarity-based \(\mathcal{L}{apn}\) achieves an average mAP of 82.4\% and an average R1 of 75.5\%, both of which are excellent results.
Furthermore, using our designed \(\mathcal{L}{apnce}\), which combines \(\mathcal{L}{apn}\) and \(\mathcal{L}{i2tce}\), results in a 0.5\% improvement in mAP compared to using only \(\mathcal{L}{i2tce}\).

Throughout the entire training process, spanning three stages, we conduct in-depth analyses of the impact of various hyperparameters.
In the first stage, we primarily assess the influence of the initialization epoch on the experimental results, as shown in Fig.~\ref{fig:4}.
The figure illustrates the performance trends over 10 initial epochs.
Compared to the baseline, mAP and R1 exhibit a rapid increase when adding only one additional epoch for the initialization training.
When the initialization epochs are increased to 3, the mAP peaks at 82.1\%, but gradually decreases, reaching 80.7\% at 10 epochs.
Upon analysis, it is speculated that a small number of epochs in the initial stage may endow the model with the ability to familiarize itself with the task but might fall short in capturing finer-grained domain-relevant information.
However, as the number of epochs increases, the capabilities of the image-encoder also enhance, potentially leading to a decline in generalization performance.

\textbf{Visualization.}
To comprehensively illustrate effectiveness of our method, we visualize t-distributed Stochastic Neighbor Embedding (t-SNE)~\cite{Maaten_Hinton_2008} on the unseen target domain dataset under Protocol-1.
As shown in Fig.~\ref{fig:5}, subplots Fig.~\ref{fig:5a} and Fig.~\ref{fig:5b} show that more dispersed and overlapping scatter plots across different domains indicate less overfitting to domain-relevant features, it is evident that our method yields embedding scatter points with a smaller domain gap compared to the baseline.
This indicates that our method effectively mitigates the issue of domain generalization.
Secondly, subplots Fig.~\ref{fig:5c}, Fig.~\ref{fig:5d}, Fig.~\ref{fig:5e} and Fig.~\ref{fig:5f} represent the Query and Gallery relationship: more overlap indicates better retrieval performance.
Thus, as observed, our method demonstrates significantly more overlap between query and gallery scatter points across various datasets compared to the baseline, indicating notably improved retrieval accuracy.

\begin{figure}[h]

  \begin{subfigure}{0.48\linewidth}
    \includegraphics[width=\linewidth]{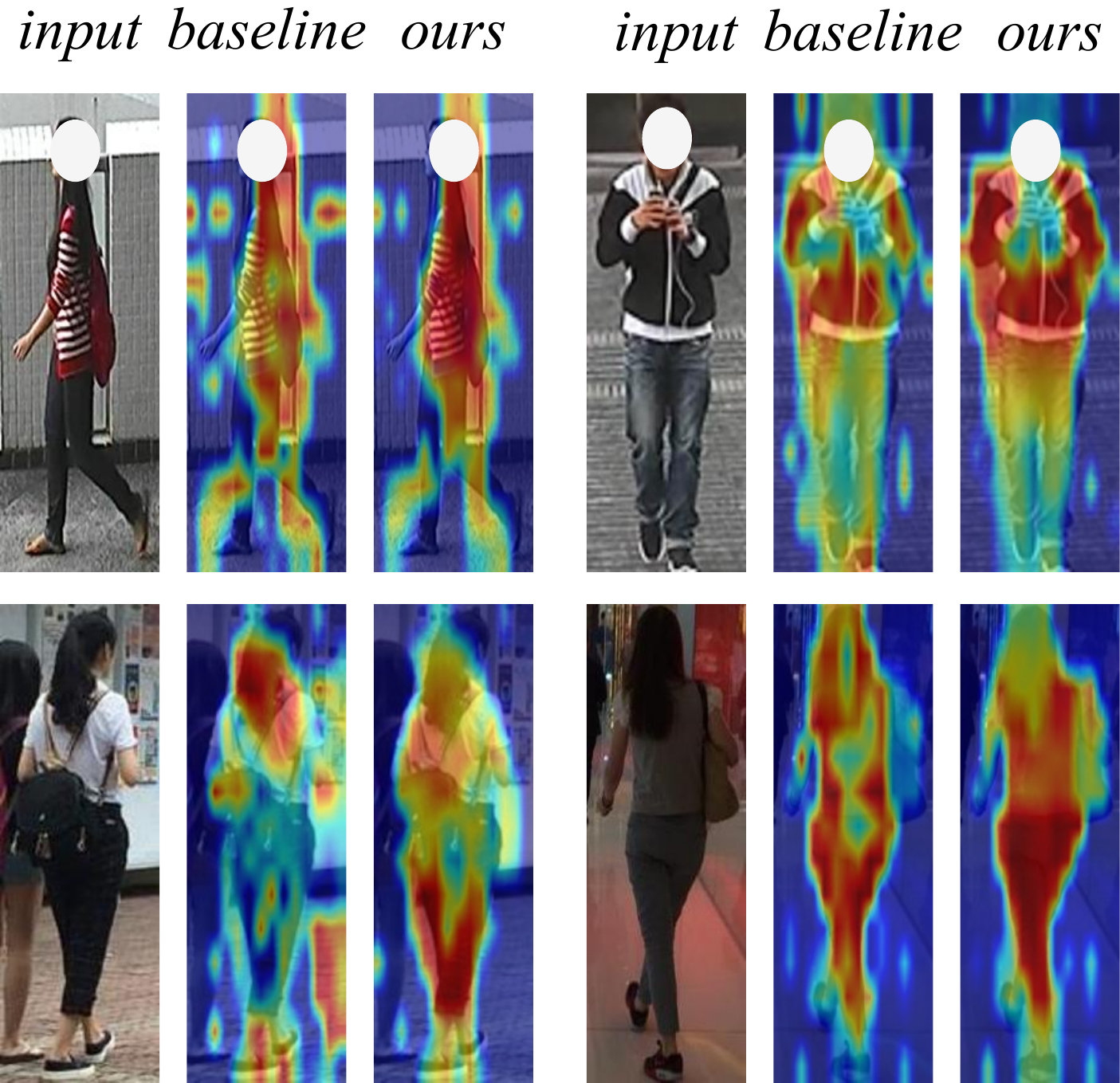}
    \caption{Source domain}
    \label{fig:sub1}
  \end{subfigure}%
  \hfill
  \begin{subfigure}{0.48\linewidth}
    \includegraphics[width=\linewidth]{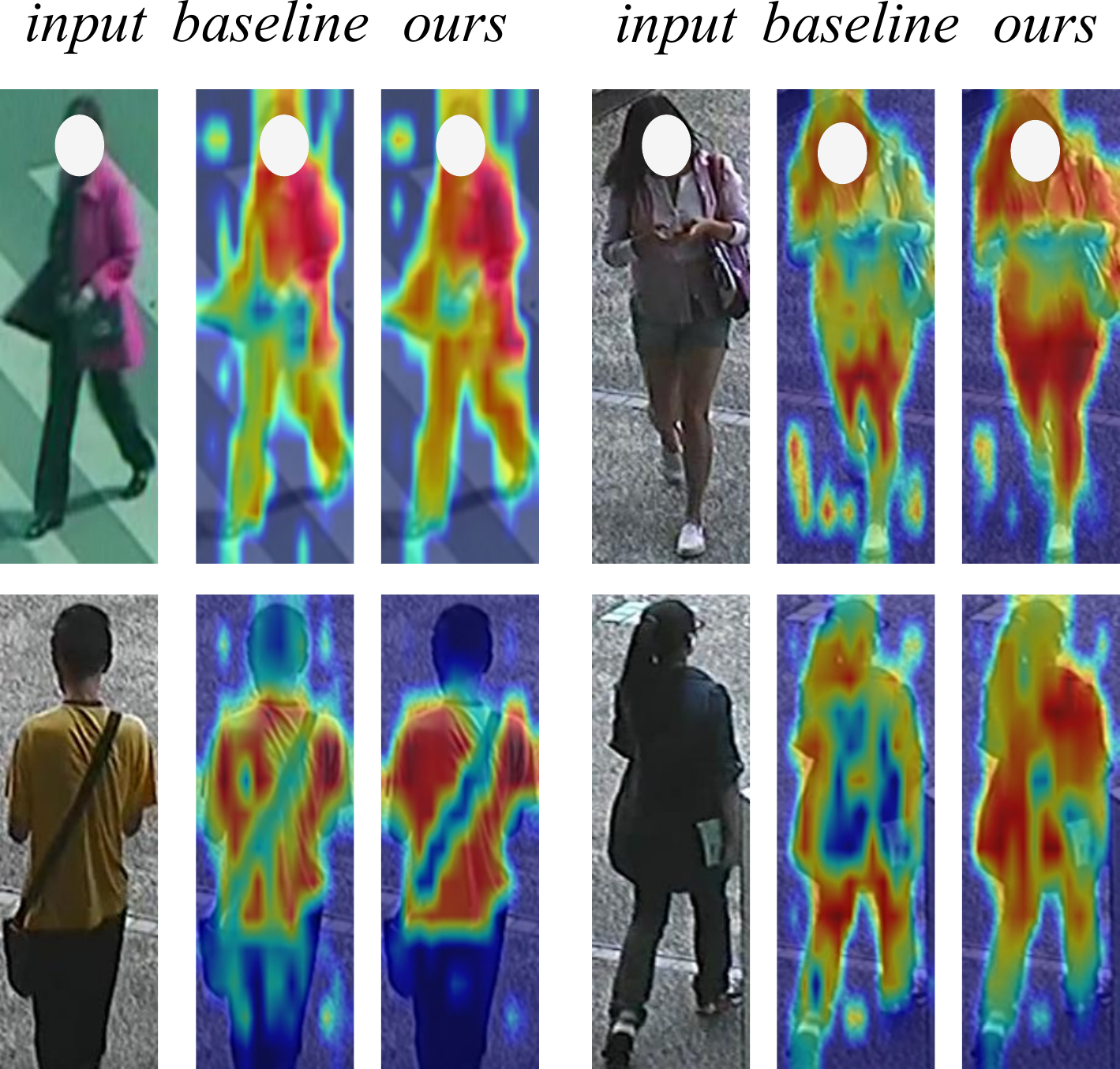}
    \caption{Target domain}
    \label{fig:sub2}
  \end{subfigure}

  \caption{Activation maps. (a) and (b) come from the source domain and the target domain respectively, the first column is the original image, the second column is the activation map from the baseline and the third column is the activation map from our method.}
    \label{fig:cam}
     \vspace {-1.2em}
\end{figure}

In addition to this, we reference GradCAM~\cite{selvaraju2017grad} to draw the attention map, as shown in Fig.~\ref{fig:cam}.
The first column is the original image, the second column is the activation map from the baseline, and the third column is the activation map from our method.
Moreover the first row represents the front view of a person, and the second row represents the back view.
Regardless of whether it is the front or the back, our method pays less attention to the environment and focuses more on the person itself compared to the baseline.
Fig.~\ref{fig:sub1} shows the attention map obtained from the training set, and Fig.~\ref{fig:sub2} shows the attention map for the test set (unseen domain).
And it can be seen that our method achieves good results in both the source domain and the target domain.

\section{Conclusion}
\label{sec:conclusion}

In conclusion, our study demonstrated that vision-language models like CLIP can be highly effective in representing visual features for generalizable person re-identification.
The proposed three-stage strategy, which incorporates a bidirectional guiding method, diligently works in each stage to learn fine-grained yet domain-invariant features.
This comprehensive method not only highlights the powerful capabilities of CLIP in learning fine-grained features but also ensures generalization across various domains.
The results of our experiments underscore the model's excellence, achieving outstanding performance and validating the effectiveness of our proposed method in enhancing the discriminative power and generalization ability of person re-identification.

\textbf{Limitation.}
Upon examining our methodology, we affirm its effectiveness and value.
However, a potential limitation is the complexity of our three-stage learning paradigm compared to conventional single-stage approaches.
While this complexity enhances feature learning, it may challenge simplicity and ease of implementation, warranting further study.
Additionally, although our method primarily relies on the CLIP model, we explore its application to other VLMs like BLIP.
Implementing our method on newer VLMs may present challenges, making this a valuable direction for future research.


%
%

\ifCLASSOPTIONcaptionsoff
  \newpage
\fi

\ifCLASSOPTIONcaptionsoff
  \newpage
\fi
\bibliographystyle{IEEEtran}
\bibliography{main}

 \vfill

\end{document}